\documentclass[final,leqno,onefignum,onetabnum]{siamart190516}

\usepackage{amsmath,amssymb}
\usepackage{supertabular,multirow}
\usepackage{graphicx}
\usepackage{epsfig}
\usepackage{subfigure}
\usepackage{bm}
\usepackage{color}
\usepackage{algorithm}
\usepackage{algorithmic}
\usepackage{appendix}

\title{Generative Adversarial Network for Probabilistic Forecast of Random Dynamical System} 

\author{Kyongmin Yeo\thanks{IBM T.J. Watson Research Center, Yorktown Heights, NY, USA  (\email{kyeo@us.ibm.com}). }
            \and Zan Li\thanks{Electrical, Computer, and Systems Engineering Department, Rensselaer Polytechnic Institute, Troy, NY, USA}
            \and Wesley M.~Gifford\footnotemark[1]}

\begin{document}

\maketitle
	\slugger{sisc}{xxxx}{xx}{x}{x--x}

\begin{abstract}
We present a deep learning model for data-driven simulations of random dynamical systems without a distributional assumption. The deep learning model consists of a recurrent neural network, which aims to learn the time marching structure, and a generative adversarial network  (GAN) to learn and sample from the probability distribution of the random dynamical system. Although GANs provide a powerful tool to model a complex probability distribution, the training often fails without a proper regularization. Here, we propose a regularization strategy for a GAN based on consistency conditions for the sequential inference problems. First, the maximum mean discrepancy (MMD) is used to enforce the consistency between conditional and marginal distributions of a stochastic process. Then, the marginal distributions of the multiple-step predictions are regularized by using MMD or from multiple discriminators. The behavior of the proposed model is studied by using three stochastic processes with complex noise structures.
\end{abstract}
	
	

\pagestyle{myheadings}
\thispagestyle{plain}

\section{Introduction}
Learning the temporal evolution of a complex dynamical system from noisy time series data is a long-standing research topic across many disciplines, including statistics, physics, engineering, and computer science \cite{Box08,Lim21,Mudelsee19,Wang16}. While there has been much progress in time series models for nonlinear dynamical systems with at least partial knowledge on the underlying time marching structure \cite{Arula02,Wiljes19,Evensen03,Hamilton15}, modeling of a nonlinear dynamical system without prior knowledge of the governing equations still remains as a challenging topic. Recently, deep learning approaches have shown promising results in the data-driven learning of complex dynamics due to the strong capability of artificial neural networks in learning nonlinear structures in the data \cite{Champion19,Genty21,Pathak18,Yeo19}. There are two major deep-learning approaches in the data-driven modeling of dynamical systems. In the first approach, deep learning models are developed to approximate / discover the time marching operator of a dynamical system from a time series data \cite{Champion19,Lu21,Lusch18,Qin19}.  On the other hand, the second approach focuses on directly modeling the time series data by a variant of recurrent neural network (RNN) \cite{Agarwal21,Pathak18,Vlachas18,Yeo21}. Although the previous studies demonstrate the potential of deep learning in learning complex dynamical systems, data-driven modeling of a \emph{random} dynamical system remains a challenging topic, particularly when the time marching operator exhibits time-delay dynamics or the noise process has a complex probability distribution. In this study, we consider the second class of deep learning approaches; we aim to develop a noble RNN model for inference and simulation of time series data from a random dynamical system without a prior knowledge on the system dynamics. 

Let $\bm{x}_t \in \mathbb{R}^d$ denote a $d$-dimensional stochastic process at time step $t$. In the time series inference problem, we aim to find the data generating distribution, $p(\bm{x}_1,\cdots,\bm{x}_T)$, from the time series data of length $T$. As discussed in  \cite{GoodfellowBengio16}, a RNN is a nonlinear extension of the state-space model. Using a RNN, the challenging inference problem of learning the full joint probability distribution, $p(\bm{x}_1,\cdots,\bm{x}_T)$, can be converted into a set of simpler inference problems of estimating $p(\bm{x}_t|\bm{h}_t)$ for $t=1,\cdots,T$, in which $\bm{h}_t$ is the hidden state of the RNN \cite{Yeo21}. Since the hidden state of a RNN is a function of the past trajectory, \emph{i.e.}, $\bm{h}_t = f(\bm{x}_1,\cdots,\bm{x}_{t-1})$, a RNN is capable of capturing the time-delay dynamics \cite{Gers00}. Although RNNs have shown outstanding performance in learning the long-time dependence and nonlinear time-marching operator from the data, it still requires an assumption on the distributional property of the random variable, $\bm{x}_t$. Usually, a multivariate Gaussian distribution with a diagonal covariance is assumed, \emph{i.e.},
\[
p(\bm{x}_t|\bm{h}_t) = \prod_{i=1}^d p(x_{i_t}|\bm{h}_t) = \prod_{i=1}^d \mathcal{N}(x_{i_t};\mu_i(\bm{h}_t),\sigma^2_i(\bm{h}_t)),
\]
in which $\mu_i(\cdot)$ and $\sigma_i(\cdot)$ denote the functions for the mean and standard deviation of the Gaussian distribution \cite{chung15,Salinas20}. This assumption on the probability distribution of $\bm{x}_t$ significantly restricts the capability of RNNs, as it only captures a unimodal distribution and also makes an independence assumption. To relax these restrictions, \cite{Yeo19} developed a RNN model to directly approximate the probability distribution using numerical discretization, and \cite{Salinas19} proposed to use a low-dimensional Gaussian Copula to approximate the joint probability distribution. 

Since its introduction \cite{Goodfellow14}, the Generative Adversarial Network (GAN) has attracted great attention due to its strength in (implicitly) learning a complex probability distribution \cite{Goodfellow20,Hong19,Saxena21}. Recently, GAN and its variants have been successfully applied to a wide range of physics problems \cite{Ahmed21,Kim21,Otten21,Paganini18,Yang20}. In contrast, GANs for sequential time series inference of noisy observations of complex dynamical systems are not relatively well studied. While there are numerous studies about GANs for time series prediction \cite{Wu20,Yoon19,Yu17}, most of the previous studies focus on generating ``realistic'' looking time series or na\"{i}ve applications of GANs to time series data. Hence, the behaviors of GANs in the data-driven predictions of a stochastic process are not well understood.

Here, we aim to develop a more general deep learning framework to learn a complex multi-dimensional joint probability distribution of a stochastic process, $\bm{x}_t$, without a distributional assumption and to simulate the learned stochastic dynamics to predict the probability distribution of the future system state. Here, we propose to use a RNN to learn the time evolution of the stochastic process and a GAN to learn and draw samples from the conditional distribution, $p(\bm{x}_{t+1}|\bm{x}_{1:t})$, in which $\bm{x}_{a:b} = (\bm{x}_a,\cdots,\bm{x}_b)$. Although a GAN does not explicitly provide the probability distribution, since a Monte Carlo simulation is used to evaluate the future system state of a nonlinear stochastic process, it is enough to be able to draw a sample from the probability distribution of the stochastic process. We propose regularization strategies based on the consistency between conditional and marginal distributions of the time series problem. This paper is organized as follows; the problem set up is described in section \ref{sec:Prob}. Then, a brief review of GAN is given in section \ref{sec:GAN}. Section \ref{sec:CR-GAN} describes our GAN approach to learn and simulate a random dynamical system. The numerical experiments of our GAN model are showed in section \ref{sec:results}. Finally, the conclusions are given in section \ref{sec:summary}.

\section{Methodology}\label{sec:numerics}

\subsection{Problem formulation}\label{sec:Prob}

Let $\bm{x}(t) \in \mathbb{R}^d$ denote the state of a physical system.  The time evolution of $\bm{x}(t)$ is governed by the following stochastic differential equation
\begin{equation}
\frac{d\bm{x}(t)}{dt} = \mathcal{F}(\bm{x}(t),\bm{x}(t-\tau)) + \bm{\epsilon}(t),
\end{equation}
in which $\bm{\epsilon}(t)$ is a noise process, $\tau$ is a delay time, and $\mathcal{F}(\cdot)$ is a deterministic time marching operator. Here, we assume no prior knowledge about the system, \emph{i.e.}, $\mathcal{F}$, $\tau$, and $\bm{\epsilon}$. The time series data set ($\bm{\mathcal{X}}$) is generated by sampling $\bm{x}(t)$ with a fixed sampling interval, $\delta t$: $\bm{\mathcal{X}} =\{ \bm{x}_0,\cdots,\bm{x}_T \}$, where $\bm{x}_t = \bm{x}(t_0 + t \delta t)$ and $t_0$ is the sampling start time.

We consider learning conditional probability density functions, $p(\bm{x}_t|\bm{x}_{0:t-1})$ for $t\in[1,T]$, from $\bm{\mathcal{X}}$ by using a recurrent neural network supplemented by GAN. Once the conditional probability distribution is learned, the simulation is performed by successively computing the conditional probability distribution functions,
\[
p(\bm{x}_{t+n},\cdots,\bm{x}_{t+1}|\bm{x}_{0:t}) = \prod_{i=1}^n p(\bm{x}_{t+i}|\bm{x}_{0:t-1},\bm{x}_{t:t+i-1}).
\]
The future joint probability distribution is evaluated using Monte Carlo simulation, and we use a GAN to draw the samples from the conditional distribution.

\subsection{Review of the Generative Adversarial Network} \label{sec:GAN}

A Generative Adversarial Network is trained by the mini-max game between two functions, discriminator and generator, to find a Nash equilibrium \cite{Goodfellow20}. The discriminator, $D: \mathbb{R}^m \rightarrow (0,1)$, computes the probability of the input $\bm{y} \in \mathbb{R}^m$ being a sample from the data set, and the generator, $G:\mathbb{R}^n \rightarrow \mathbb{R}^m$, outputs $\bm{y}^* \in \mathbb{R}^m$, from an input, $\bm{z} \in \mathbb{R}^n$. Typically, artificial neural networks are used to parameterize both $D(\bm{y})$ and $G(\bm{z})$. The input to the generator, $\bm{z}$, is a random variable with a simple probability distribution, such as an isotropic Gaussian distribution. The optimization problem to train GAN is
\begin{equation} \label{eqn:GAN_loss}
\min_{G}\max_D E_{\bm{y}\sim p_d} \left[ \log D(\bm{y}) \right] + E_{\bm{z} \sim p_z} \left[  \log \left(1-D(G(\bm{z}))\right) \right],
\end{equation}
in which $p_d(\bm{y})$ is the data generating distribution and $p_z(\bm{z})$ is the probability distribution of $\bm{z}$. The mini-max optimization problem aims to make the discriminator better at distinguishing the true samples (from data) from the fake samples (from $G(\bm{z})$) and, at the same time, to make $G(\bm{z})$ generate outputs closer to the true samples so that $D(\bm{y})$ fails to distinguish between true and fake samples. It was shown that at the global minimum, $p_d(\bm{y}) =p_g(\bm{y})$, in which $p_g(\bm{y})$ is the probability distribution of the generated samples \cite{Goodfellow14}. Theoretical analysis of GAN is an active area of research \cite{Fedus18,Heusel17,Nagarajan17,Sun20_NIPS}. 
Since the majority of the applications of GANs are either in image or text generation, it is typical to use ``real'' and ``fake'' to explain how GANs work, \emph{e.g.}, the discriminator distinguishes between ``real'' and ``fake'' samples. However, when $\bm{y}$ is  a real-valued random variable, the discussions based on ``real'' and ``fake'' samples are not straightforward. Here, we provide a simpler analysis of GAN for a real-valued random variable, $\bm{y}$.

Without loss of generality, we consider a one-dimensional random variable $y \in \mathbb{R}$. While the true data generating distribution, $p_d$, may not have a compact support, we can identify the range of a data set, $\mathcal{Y} = \{ y_1,\cdots,y_N \}$, in which $N$ is the total number of data,
\[
Supp(p_d) = [y_{min},y_{max}].
\]
Here, $y_{min}$ and $y_{max}$ denote the minimum and maximum of $\mathcal{Y}$, respectively, and, with a slight abuse of notation, we use $Supp(p_d)$ to denote the range of the data. We can define a set of ordered real numbers, $\bm{\alpha} = \{ \alpha_1,\cdots,\alpha_{K-1} \}$, such that
\[
y_{min} < \alpha_1 < \cdots < \alpha_{K-1} < y_{max},
\] 
and associated partitions, $\mathcal{A} = \{A_i; A_i = [\alpha_{j-1},\alpha_j), i\in\mathbb{N}_{>0}, i \le K\}$, where $\alpha_0 = y_{min}$ and $\alpha_K = y_{max}$. 
Define a mapping $\mathcal{C}:\mathbb{R}\rightarrow\mathbb{N}_{>0}$, such that
\[
\mathcal{C}(y) = i,~\text{when}~y \in A_i. 
\]
Then, we can define a discrete random variable, $Y = \mathcal{C}(y)$, and a discrete probability distribution, $P_i = p(Y=i)$. The probability distribution can be calculated as,
\[
P_i = \frac{1}{N} \sum_{j=1}^N \chi_{A_i}(y_j),
\]
where $\chi_{A_i}(y)$ is an indicator function that returns one if $y \in A_i$ and zero otherwise. 

Similarly, we can define a discrete probability distribution of the generated samples. Let $z \in \mathbb{R}$ be a random variable with the probability distribution, $p_z(z)$. The discrete probability distribution of the \emph{generated samples} can be defined as $Q_i = p(Y^* = i)$, in which $Y^* = \mathcal{C}(G(z))$. Similar to $P_i$, $Q_i$ is evaluated as
\[
Q_i = \frac{1}{M} \sum_{j=1}^M \chi_{A_i}(G(z_j)),
\]
where $M$ is the number of realizations.
Now, the optimization problem of the GAN (\ref{eqn:GAN_loss}) is simplified as
\begin{equation}\label{eqn:discrete_opt}
 \min_{\bm{Q}}\max_{\bm{D}} \sum_{i=1}^K P_i \log D_i + Q_i \log (1-D_i),
\end{equation}
in which $D_i$ is a discriminator function for the partition $A_i$. Note that in (\ref{eqn:discrete_opt}) the minimization is over the probability distribution of the generated samples, $\bm{Q}$.

In the training of a GAN, usually a stochastic gradient descent method is used alternating between two steps; first update $D$ for $G$ fixed (D-Step) and update $G$ for $D$ fixed (G-Step). The optimization problem of D-Step, 
\begin{equation}
\max_{\bm{D}} \sum_{i=1}^K P_i \log D_i + Q_i \log (1-D_i),
\end{equation}
has a simple solution,
\begin{equation}
D_i = \frac{1}{1+Q_i/P_i}~\text{for}~1\le i \le K.
\end{equation}
It suggests that the discriminator score is based on the relative frequency of the samples observed in each partition. If we assume $M=N$, $D_i$ becomes larger than 0.5 when we have more data than generated samples in $A_i$, \emph{i.e.}, $P_i > Q_i$, and vice versa. On the other hand, in G-Step, the optimization problem is
\begin{equation} \label{eqn:g_step}
\min_{\bm{Q}} \sum_{i=1}^K Q_i \log(1-D_i)~\text{subject to}~\sum_{i=1}^K Q_i = 1,
\end{equation}
of which solution is simply $Q_i = \delta_{ik}$ for $k = \max_{j} D_j$ when $\bm{D}$ has only one maximum. If $\bm{D}$ has multiple maxima, $\bm{Q}$ cannot be uniquely determined. When D-Step and G-Step are considered simultaneously, the equilibrium solution is given as
\begin{equation}
D_1=\cdots=D_K = 0.5,~\text{and}~P_i = Q_i~\text{for}~1 \le i \le K.
\end{equation}
However, in practice, it is very difficult to achieve this \emph{Nash equilibrium} in the training \cite{Farnia20,Goodfellow20}.

It is important to note that, in the analysis above, we implicitly assumed that $\bm{P}$ and $\bm{Q}$ have the same support. Let $Supp(p_g)$ denote the support of $\bm{Q}$. If $Supp(p_g) \setminus Supp(p_d) \neq \emptyset$, $\bm{Q}$ in $Supp(p_g) \setminus Supp(p_d)$ will be penalized by $\bm{D}$, which will shrink $Supp(p_g)$. On the other hand, when $Supp(p_d) \setminus Supp(p_g) \neq \emptyset$, because the optimization problem of G-Step (\ref{eqn:g_step}) is defined only within $Supp(p_g)$, there is no mechanism to make $Supp(p_g)$ bigger to cover $Supp(p_d)$. When this happens, the optimization problem fails to converge.

There are two major concerns in training a GAN. First, because the discriminator score at a point $\bm{y}$ depends on the number of data around $\bm{y}$, training of a GAN suffers from the curse of dimensionality. As the dimension of $\bm{y}$ increases, the number of data required to uniformly cover the space increases exponentially. Second, while it is required to have $Supp(p_d) = Supp(p_g)$ for the training of a GAN to converge, it is difficult to achieve because the optimization in G-Step is defined only on $Supp(p_g)$, without considering $Supp(p_d)$. Because of these issues, training a GAN often fails. To overcome such difficulties, many efforts have been focused on regularizing the behavior of the discriminator \cite{Fedus18,Roth17,Thanh-Tung19,Zhang20}. On the other hand, regularizing the generator has not been widely explored.

\subsection{Consistency-Regularized Generative Adversarial Network} \label{sec:CR-GAN}

\subsubsection{Generative Adversarial Network}

\begin{figure}
	\center \includegraphics[width=0.9\textwidth]{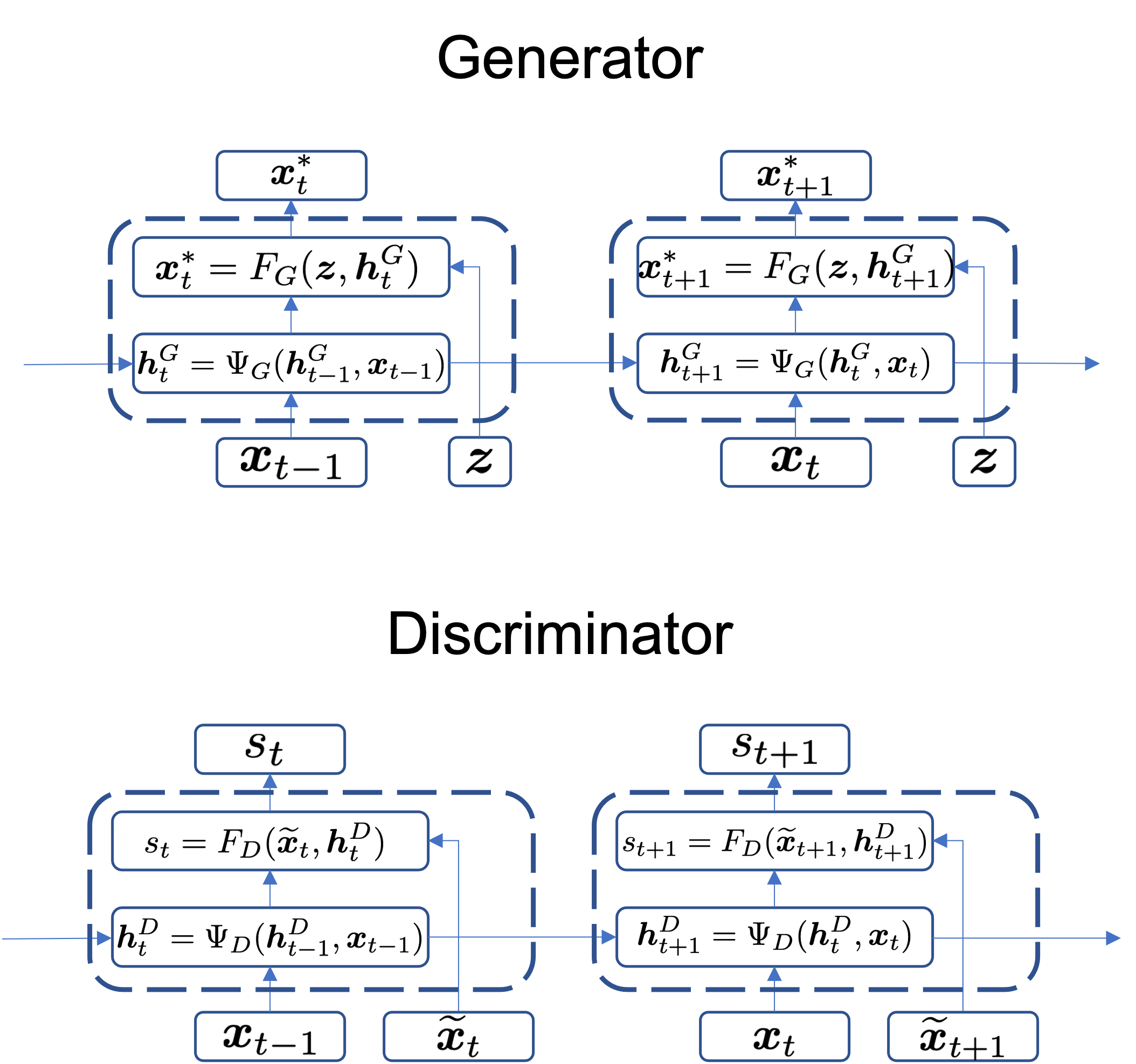}
	\caption{Sketches of the Generator (left) and the Discriminator (right). $\bm{x}^*_t$ indicates a sample drawn from $p_g(\bm{x}_t|\bm{x}_{0:t-1})$, and $s_t$ denotes the discriminator score at time step $t$. In the Generator, a realization of the random variable, $\bm{z}$, is fed into $F_G$ together with the updated internal state of the generator RNN, $\bm{h}^G_{t+1}$, to generate a sample for the next time step. In the Discriminator, first the discriminator RNN is updated with the current value of the ground truth trajectory, e.g., $\bm{x}_t$. Then, a sample for the next time step, $\widetilde{\bm{x}}_{t+1}$, is fed into $F_D$ together with the updated internal state, $\bm{h}^D_{t+1}$, where the Discriminator computes the discriminator score of $\widetilde{\bm{x}}_{t+1}$. The sample, $\widetilde{\bm{x}}_{t+1}$, may be from the data, $\bm{x}_{t+1}$, or from the generator, $\bm{x}^*_{t+1}$.} \label{fig:sketch}
\end{figure}

Here, we aim to learn the conditional probability distribution, $p_d(\bm{x}_{t+1}|\bm{x}_{0:t})$, of a $d$-dimensional stochastic process $\bm{x}_t\in\mathbb{R}^d$ for the data-driven simulation. Since the fully connected graph structure of the conditional distribution, \emph{i.e.}, the dependence on the entire historical sequence, $\bm{x}_{0:t}$, makes the inference problem difficult, we use a RNN and exploit the conditional independence, such that
\begin{equation}
p_g(\bm{x}_{t+1}|\bm{x}_{0:t}) = p_g(\bm{x}_{t+1}|\bm{h}^G_{t+1}).
\end{equation}
Here, $\bm{h}^G_t$ is the hidden state of the RNN. The hidden state is updated as
\begin{equation}
\bm{h}^G_{t+1} = \Psi_G(\bm{x}_t,\bm{h}^G_t),
\end{equation}
in which $\Psi_G(\cdot)$ denotes a nonlinear transformation defined by the RNN.  On top of the RNN, we add a feedforward network, $F_G(\bm{z},\bm{h}^G_{t+1})$, which aims to draw a sample from $p_g(\bm{x}_{t+1}|\bm{x}_{0:t})$ by using a random variable, $\bm{z}\in\mathbb{R}^n$, as an input variable. We use an isotropic Gaussian distribution for $\bm{z}$; $\bm{z} \sim \mathcal{N}(\bm{0},\bm{I})$. A sketch of the generative model, or generator $G(\bm{x}_t,\bm{h}^G_t,\bm{z})$, is shown in the left panel of Fig. \ref{fig:sketch}.

As discussed in section \ref{sec:GAN}, given a sample at time $t+1$, $\widetilde{\bm{x}}_{t+1}$, the discriminator aims to approximate the ratio between the data and generator distributions at $\widetilde{\bm{x}}_{t+1}$ given the historical sequence, $\bm{x}_{0:t}$, \emph{i.e.}, $p_g(\widetilde{\bm{x}}_{t+1}|\bm{x}_{0:t}) / p_d(\widetilde{\bm{x}}_{t+1}|\bm{x}_{0:t})$. Hence, we again use a RNN, $\Psi_D(\cdot)$, and its associated hidden state, $\bm{h}^D_t$, 
\begin{equation}
\bm{h}^D_{t} = \Psi_D(\bm{x}_{t-1},\bm{h}^D_{t-1}).
\end{equation}
Then, a feedforward neural network, $F_D(\widetilde{\bm{x}}_t,\bm{h}^D_t)$, is added to compute the discriminator score, $s_t$. The right panel of Fig. \ref{fig:sketch} shows a sketch of the discriminator.

When training a GAN for a time series problem, it is typical to follow the original GAN training approach \cite{Wu20,Yoon19,Yu17}. In other words, first compute the first term of the objective function in (\ref{eqn:GAN_loss}) by using the data $\bm{x}_{0:t}$, and then generate a sequence, $\bm{x}^*_{0:t}$ from $\bm{z}$ to compute the second term of (\ref{eqn:GAN_loss}) from $\bm{x}^*_{0:t}$. However, in this approach, the optimization problem  of the time series problem becomes
\begin{align}
\min_{G}\max_{D}~  E_{\bm{x}_{0:t} } [ \log D(\bm{x}_{0:t}) ] + E_{\bm{x}^*_{0:t} } [ \log (1-D(\bm{x}^*_{0:t})) ],
\end{align}
in which $E_{\bm{x}}[\cdot]$ and $E_{\bm{x}^*}[\cdot]$ denote expectations over the data distribution, $\bm{x} \sim p_d$, and the generator distribution, $\bm{x}^* \sim p_g$, respectively. Note that the generator in the second term is replaced by the expectation over $p_g$. In this formulation, the discriminator tries to learn the ratio between the full joint probability distribution, $p_g(\bm{x}_{0:t})/p_d(\bm{x}_{0:t})$. As shown in section \ref{sec:GAN}, learning the full joint probability distribution is challenging due to the high dimensionality. 

Here, we propose to use the following objective function to learn the conditional distribution, not the full joint probability distribution. At time step $t$, we propose to train a GAN using the following objective function,
\begin{equation} \label{eqn:GAN_loss_new}
 \min_{G}\max_{D}~ E_{\bm{x}_{0:t-1}} [\mathcal{L}(\bm{x}_{0:t-1})],
\end{equation}
in which
\begin{equation} \label{eqn:GAN_loss_new_2}
\mathcal{L}(\bm{x}_{0:t-1}) = \int p_d(\bm{x}_t|\bm{x}_{0:t-1}) \log D(\bm{x}_t|\bm{x}_{0:t-1}) +p_g(\bm{x}_t|\bm{x}_{0:t-1}) \log (1- D(\bm{x}_t|\bm{x}_{0:t-1})) d\bm{x}_t.
\end{equation}
Note that the GAN loss (\ref{eqn:GAN_loss_new_2}) is evaluated over the expectation of the data generating distribution, $p_d(\bm{x}_{0:t-1})$ as denoted in (\ref{eqn:GAN_loss_new}). The major difference from the previous methods is, instead of evaluating the second term by a sequence of generated samples as in (\ref{eqn:GAN_loss}), we use the data $\bm{x}_{0:t-1}$, draw a sample $\bm{x}^*_t$ by using the generator, and evaluate the discriminator with $\bm{x}^*_t$. Now, the objective of the discriminator at time step $t$ becomes comparing the conditional distributions, $p_g(\bm{x}_t|\bm{x}_{0:t-1})/p_d(\bm{x}_t|\bm{x}_{0:t-1})$. In other words, the discriminator is used to distinguish between the data, $\bm{x}_t$, and a generated sample, $\bm{x}^*_t$, conditioned on a sequence from the data, $\bm{x}_{0:t-1}$, instead of comparing a sequence, $\bm{x}_{0:t}$, to a sequence, $\bm{x}^*_{0:t}$. 

\subsubsection{Consistency Regularization}
In this section, we propose regularization strategies to regularize the behaviors of the generator to overcome the difficulties in the training of GANs discussed in section \ref{sec:GAN}. In the time series problem, there can be multiple ways of representing the probability distribution of the data. While we are interested in learning the conditional distribution, $p(\bm{x}_t|\bm{x}_{0:t-1})$, the conditional distribution should be consistent with the marginal distribution, $p(\bm{x}_t)$, through an integral constraint;
\begin{equation}
\int p(\bm{x}_t|\bm{x}_{0:t-1}) p(\bm{x}_{0:t-1}) d\bm{x}_{0:t-1} = p(\bm{x}_t).
\end{equation}
Based on this consistency condition, we aim to regularize the generator such that the marginal distribution of the generated samples, $p_g(\bm{x}_t)$, becomes closer to $p_d(\bm{x}_t)$, \emph{i.e.},
\begin{equation} \label{eqn:cond_marginal}
\int p_g(\bm{x}_t|\bm{x}_{0:t-1}) p_d(\bm{x}_{0:t-1}) d\bm{x}_{0:t-1} \simeq p_d(\bm{x}_t).
\end{equation}
However, since we do not know $ p_g(\bm{x}_t|\bm{x}_{0:t-1})$ and $p_d(\bm{x}_t)$ explicitly, we cannot directly evaluate (\ref{eqn:cond_marginal}). Here, we use the maximum mean discrepancy (MMD) to compare the left and right hand sides of (\ref{eqn:cond_marginal}) from the samples drawn from the data and the generator. The MMD is a two-sample test to compare two distributions in a reproducing kernel Hilbert space \cite{Gretton12}. Using the Gaussian kernel, MMD for the samples from two distributions, $\bm{x} \sim p_d(\bm{x})$ and $\bm{x}^* \sim p_g(\bm{x})$, is computed as
\begin{equation}
\text{MMD}(\bm{x},\bm{x}^*) = \frac{1}{T^2} \left[ \sum_{i=1}^T\sum_{j=1}^T k(\bm{x}_i,\bm{x}_j) + \sum_{i=1}^T\sum_{j=1}^T k(\bm{x}^*_i,\bm{x}^*_j) - 2\sum_{i=1}^T\sum_{j=1}^T k(\bm{x}_i,\bm{x}^*_j) \right],
\end{equation}
in which
\begin{equation} \label{eqn:mmd_kernel}
k(\bm{x},\bm{x}^*) = \exp \left( -\frac{1}{\gamma} \|\bm{x} - \bm{x}^*\|^2_2 \right).
\end{equation}
Here, we assume that the sizes of the two samples, $\bm{x}$ and $\bm{x}^*$, are the same ($T$).

With the MMD regularization, the optimization problem becomes
\begin{equation} \label{eqn:CR-GAN_loss_1}
\min_{G}\max_{D} \left\{ \sum_{t=1}^T E_{\bm{x}_{0:t-1} } [\mathcal{L}(\bm{x}_{0:t-1})]\right\} + \lambda \mathop{E}_{\substack{\bm{x} \sim p_d \\ \bm{x}^* \sim p_g}} [ \text{MMD}(\bm{x},\bm{x}^* )],
\end{equation}
in which $\lambda$ is a regularization coefficient. In this formulation, the modified GAN objective function makes the generator learn the temporal evolution of the underlying probability distribution, while the MMD regularization enforces the marginal distribution of the generated samples to be similar to the marginal distribution of the data. 

Now, we introduce another consistency condition based on the predictive distributions of a multiple-step prediction. In this study, we are interested in predicting the future system state as,
\begin{equation}
\bm{x}_{t+1},\cdots,\bm{x}_{t+H} \sim p(\bm{x}_{t+1},\cdots,\bm{x}_{t+H}|\bm{x}_{0:t}),
\end{equation}
where $H$ is the prediction horizon. Instead of directly drawing a sample from the future joint probability distribution, the samples are sequentially drawn from the condition distributions,
\begin{align}
\bm{x}^*_{t+1} &\sim p(\bm{x}_{t+1}|\bm{x}_{0:t}), \nonumber \\
\bm{x}^*_{t+2} &\sim p(\bm{x}_{t+2}|\bm{x}_{0:t},\bm{x}^*_{t+1}), \nonumber \\
&\vdots \nonumber \\
\bm{x}^*_{t+H} &\sim p(\bm{x}_{t+H}|\bm{x}_{0:t},\bm{x}^*_{t+1:t+H-1}). \nonumber
\end{align}
While this sequential sampling procedure is straightforward, there may be an accumulation of error in each sequential sampling step, which makes the predictive distribution diverge from the ground truth. To reduce such divergence, we impose a second consistency condition through a regularization on the generator. The second consistency condition is motivated to enforce
\begin{align} \label{eqn:multistep_pdf}
p_g(\bm{x}^*_{t+n}|\bm{x}_{0:t}) & = \int p_g(\bm{x}^*_{t+1}|\bm{x}_{0:t}) \prod_{i=1}^{n-1} p_g(\bm{x}^*_{t+i+1}|\bm{x}_{0:t},\bm{x}^*_{t+1:t+i}) d\bm{x}^*_{t+1:t+i}\\
&\simeq~ p_d(\bm{x}_{t+n}|\bm{x}_{0:t}). \nonumber
\end{align}
However, it is challenging to impose a regularization by directly comparing conditional distributions. Instead, we propose to regularize the marginal distributions for the temporal difference. 

Define a $n$-th order temporal difference as,
 \[
 \Delta_n \bm{x}_t = \bm{x}_{t+n} - \bm{x}_t.
 \]
 Then, we can impose a consistency condition by comparing $p_g(\Delta_n \bm{x}_t)$ and $p_d(\Delta_n \bm{x}_t)$, instead of dealing with the predictive distribution (\ref{eqn:multistep_pdf}) directly. Here, we propose two models to impose the second consistency condition. In the first method, we can simply add another MMD regularizer in the optimization formulation in (\ref{eqn:CR-GAN_loss_1});
\begin{align} \label{eqn:CR-GAN_loss_MMD}
\min_{G}\max_{D} \left\{ \sum_{t=1}^T E_{\bm{x}_{0:t-1} } [\mathcal{L}(\bm{x}_{0:t-1})]\right\} &+ \lambda_1 \mathop{E}_{\substack{\bm{x} \sim p_d \\ \bm{x}^* \sim p_g}} [ \text{MMD}(\bm{x},\bm{x}^* )] \\
&+ \lambda_2  \mathop{E}_{\substack{\Delta_n \bm{x} \sim p_d \\ \Delta_n \bm{x}^* \sim p_g}} [ \text{MMD}(\Delta_n \bm{x},\Delta_n \bm{x}^* )] . \nonumber
\end{align}
In the second approach, we can use another discriminator, $F_\Delta(\cdot)$, and add a new GAN objective function for the marginal distribution;
\begin{align} \label{eqn:CR-GAN_loss_GAN}
\min_{G}\max_{D,F_\Delta }& \left\{ \sum_{t=1}^T E_{\bm{x}_{0:t-1} } [\mathcal{L}(\bm{x}_{0:t-1})]\right\} + \lambda_1 \mathop{E}_{\substack{\bm{x} \sim p_d \\ \bm{x}^* \sim p_g}} [ \text{MMD}(\bm{x},\bm{x}^* )] \\
&+ E_{\Delta_n \bm{x}}[\log F_\Delta(\Delta_n \bm{x})] + E_{\Delta_n\bm{x}^*}[\log (1-F_\Delta(\Delta_n\bm{x}^*))]. \nonumber
\end{align}
Since $F_\Delta(\cdot)$ does not consider the temporal dependence, we can use a simple multilayer feed-forward neural network.

Hereafter, we use CR-GAN (Consistency-Regularized GAN) to denote the time series GAN model with only the first consistency condition (\ref{eqn:CR-GAN_loss_1}), and CR-GAN-M$_n$ and CR-GAN-G$_n$ to denote the GAN models with the MMD (\ref{eqn:CR-GAN_loss_MMD}) and with the marginal discriminator (\ref{eqn:CR-GAN_loss_GAN}), respectively, trained for the $n$-th order temporal difference, $\Delta_n \bm{x}_t$.

\subsubsection{Optimization}
In practice, CR-GAN is trained alternating between the discriminator and generator steps to update the weights of the discriminator ($\bm{\theta}_D$) and the generator ($\bm{\theta}_G$) by using a stochastic gradient descent method (SGD).  

\vspace{0.3cm} \paragraph{Discriminator Step} Given training data, $ \{\bm{x}_0,\cdots,\bm{x}_T\}$, the objective function is computed as
\begin{enumerate}
  \item Update the hidden states of the generator ($\Psi_G$) and discriminator RNN ($\Psi_D$) using the same data;
    \begin{align}
        \bm{h}^G_t &= \Psi_G(\bm{h}^G_{t-1},\bm{x}_{t-1}), \label{eqn:D_Step_1} \\
        \bm{h}^D_t &= \Psi_D(\bm{h}^D_{t-1},\bm{x}_{t-1}).        
    \end{align}
  \item Draw a sample from the generator;
    \begin{align}
      \bm{x}^*_t = F_G(\bm{z},\bm{h}^G_t),~\text{where}~\bm{z} \sim \mathcal{N}(\bm{0},\bm{I}).
    \end{align}
  \item Compute the objective function;
    \begin{align}
        \mathcal{L}_t = \log (F_D(\bm{x}_t,\bm{h}^D_t)) + \log (1-F_D(\bm{x}_t^*,\bm{h}^D_t))
    \end{align}
  \item Update the weights of the discriminator, $\bm{\theta}_D$, by using SGD;
    \begin{align} \label{eqn:D_Step_final}
        \bm{\theta}_D \leftarrow \bm{\theta}_D - \alpha_D\times \text{SGD}\left(  \nabla_{\bm{\theta}_D} \frac{1}{T} \sum_{t=1}^T \mathcal{L}_t \right),
    \end{align}
    in which $\alpha_D$ is the learning rate.
\end{enumerate}
    
\vspace{0.3cm} \paragraph{Generator Step}  Given training data, $ \{\bm{x}_0,\cdots,\bm{x}_T\}$, the objective function and the MMD regularization are computed as
\begin{enumerate}
  \item Repeat the steps 1 $\sim$ 2 of the Discriminator Step to update $\bm{h}^G_t$ and $\bm{h}^D_t$ and draw a sample $\bm{x}^*_t$.
  \item Compute the objective function;
      \begin{align}\label{eqn:G_Step_1}
        \mathcal{L}_t = - \log (F_D(\bm{x}_t^*,\bm{h}^D_t))
    \end{align}
  \item After all $\bm{x}^*_1,\cdots,\bm{x}^*_T$ are sampled in Step 1, compute the MMD loss;
    \begin{align}
      \mathcal{R}_{M} = \lambda_1 \text{MMD}(\bm{x}_{1:T},\bm{x}^*_{1:T}) \label{eqn:r_m}
    \end{align}
  \item Update the weights of the generator, $\bm{\theta}_G$, by using SGD;
    \begin{align}\label{eqn:G_Step_final}
        \bm{\theta}_G \leftarrow \bm{\theta}_G - \alpha_G\times \text{SGD}\left(  \nabla_{\bm{\theta}_G}\left\{ \frac{1}{T}\sum_{t=1}^T \mathcal{L}_t +\mathcal{R}_M\right\}  \right),
    \end{align}
    in which $\alpha_G$ is a learning rate.
\end{enumerate}

\vspace{0.3cm} \paragraph{Multiple-step Regularization Step}  Given training data, $ \{\bm{x}_0,\cdots,\bm{x}_T\}$, the regularization is computed by a multiple-step ahead prediction. First, choose the start time of the forecast, $0<t_f<T-n$.
\begin{enumerate}
  \item For $t = 1,\cdots,t_f$, update $\bm{h}^G_t$;
    \begin{equation} \label{eqn:CR_Step_1}
      \bm{h}^G_t = \Psi_G(\bm{h}^G_{t-1},\bm{x}_{t-1}).
    \end{equation}
  \item Draw a sample at $t_f$ from the generator;
    \begin{equation}
      \bm{x}^*_{t_f} = F_G(\bm{z},\bm{h}^G_{t_f}),~\text{where}~\bm{z} \sim \mathcal{N}(\bm{0},\bm{I}).
    \end{equation}
  \item For $t = t_f+1,\cdots,T$, generate a sequence of prediction;
    \begin{align}
      \bm{h}^G_t &= \Psi_G(\bm{h}^G_{t-1},\bm{x}^*_{t-1}), \\
      \bm{x}^*_{t} &= F_G(\bm{z},\bm{h}^G_{t}),~\text{where}~\bm{z} \sim \mathcal{N}(\bm{0},\bm{I}).  \label{eqn:CR_Step_2}
    \end{align}
  \item Prepare the time difference data sets, $\bm{X} =\{\Delta_n \bm{x}_{t_f},\cdots,\Delta_n \bm{x}_{T-n}\}$ and  $\bm{X}^* = \{\Delta_n \bm{x}^*_{t_f},\cdots,\Delta_n \bm{x}^*_{T-n}\}$
  \item (GAN) Update $\bm{\theta}_\Delta$ (weights of $F_\Delta$) and $\bm{\theta}_G$ from the GAN objective;
    \begin{align}
        \bm{\theta}_\Delta &\leftarrow \bm{\theta}_\Delta - \alpha_\Delta \times \text{SGD}\left(  \nabla_{\bm{\theta}_\Delta} \frac{1}{m} \sum_{i=1}^m \left\{ \log (F_\Delta (\bm{X}_i)) + \log(1-F_\Delta(\bm{X}_i^*)) \right\} \right), \label{eqn:CR_Step_3} \\
        \bm{\theta}_G &\leftarrow \bm{\theta}_G - \alpha_G \times \text{SGD}\left(  \nabla_{\bm{\theta}_G} \frac{1}{m} \sum_{i=1}^m \left\{ - \log(F_\Delta(\bm{X}^*_i)) \right\} \right),  \label{eqn:CR_Step_4}
    \end{align}
    where $m = T-t_f-n+1$, and $\bm{X}_i = \Delta_n\bm{x}_{t_f+i-1}$.
  \item (MMD) Update $\bm{\theta}_G$ from the MMD regularization
    \begin{align}  \label{eqn:CR_Step_5}
        \bm{\theta}_G \leftarrow \bm{\theta}_G - \alpha_G \times \text{SGD}\left( \lambda_2  \nabla_{\bm{\theta}_G} \text{MMD}(\bm{X},\bm{X}^*) \right),
    \end{align}
\end{enumerate}

Pseudocode for the training algorithm is shown in Algorithm \ref{alg:CR-GAN}. In CR-GAN-G$_n$ the generator should satisfy two discriminators simultaneously, and similarly in CR-GAN-M$_n$ the generator is regularized by two marginal distributions. In these cases, it is found empirically that having multiple generator updates for one discriminator update, \emph{i.e.}, $n_g \ge 2$ in Algorithm \ref{alg:CR-GAN}, provides a better performance. 

\begin{algorithm}[!tbh]
	\caption{Training of CR-GAN}
	\label{alg:CR-GAN}
	\textbf{Input}: 
	Time series data ($\mathcal{D}$), size of the mini-batch ($n_b$), length of the training sequence ($T$), forecast start time ($t_f$), regularization coefficients ($\lambda_1,\lambda_2$), learning rates ($\alpha_D,\alpha_G,\alpha_{\Delta}$), maximum number of SGD iteration ($k_{max}$), number of the generator sub-iterations ($n_g$)
	
	\begin{algorithmic}[1]
		\STATE Initialize the weights, $\bm{\theta}_G$, $\bm{\theta}_D$, $\bm{\theta}_\Delta$
		\STATE Prepare $\mathcal{\bm{X}}$, which consists of a set of length $T+1$ time series, $\bm{x}_{0:T}$, from $\mathcal{D}$
		\FOR{$k=1,k_{max}$}	
		\STATE Randomly sample a mini-batch from $\mathcal{\bm{X}}$; {\color{blue}$\bm{x}_{0:T}^{(1)},\cdots, \bm{x}_{0:T}^{(n_b)}$}
		\STATE Update $\bm{\theta}_D$ from the Discriminator Step (\ref{eqn:D_Step_final})
		\IF{CR-GAN-G}
		\STATE Update $\bm{\theta}_\Delta$ from the Mutiple-step Regularization Step (\ref{eqn:CR_Step_3})
		\ENDIF
		\FOR{$j=1,n_g$}
		\STATE Update $\bm{\theta}_G$ from the Generator Step (\ref{eqn:G_Step_final})
		\IF{CR-GAN-G}
		\STATE Update $\bm{\theta}_G$ from the Multiple-step Regularization Step (\ref{eqn:CR_Step_4})
		\ELSIF{CR-GAN-M}
		\STATE Update $\bm{\theta}_G$ from the Multiple-step Regularization Step (\ref{eqn:CR_Step_5})
		\ENDIF
		\ENDFOR
		\ENDFOR
	\end{algorithmic}
\end{algorithm}


\section{Numerical experiments} \label{sec:results}

In this section, the behaviors of CR-GAN are studied from numerical experiments for three data sets. The data sets are generated from an autoregressive process of order one with Bi-Gaussian noise, a noisy observation of the Mackey-Glass time series, and a random Lorenz system. The structure of the artificial neural networks is kept the same for all three experiments. We assume that the dimension of the random input, $\bm{z}_t$, is equal to the dimension of the time series data, $\bm{x}_t$, \emph{i.e.}, $\dim(\bm{z}_t) = \dim(\bm{x}_t)$. The artificial neural network is trained by using a variant of SGD, ADAM \cite{Kingma15}, with a mini-batch size of 100. The initial learning rate is set to $\alpha_D = \alpha_G = \alpha_{\Delta} = 5\times10^{-5}$, and gradually reduced to $10^{-5}$ by using the cosine learning-rate decay schedule \cite{Loshchilov17}. The total number of SGD iterations is set to $40,000$. A scaled variable, $\widehat{\bm{x}}_t$, is used for the input and output of the artificial neural network. The time series data, $\bm{x}_t \in \mathbb{R}^d$, is scaled such that
\begin{equation}
\widehat{x}_{t_i} =  \frac{x_{t_i} - \min(x_i)+\nu}{\max(x_i) - \min(x_i)+2\nu},~\text{for}~i=1,\cdots,d,
\end{equation}
in which $\max(x_i)$ and $\min(x_i)$, respectively, indicate the maximum and minimum values of the $i$-th dimension of $\bm{x}_t$, and $\nu \ge 0$ is a small number. When $\nu = 0$, $\widehat{\bm{x}}_t$ is defined in a hypercube, $\widehat{\bm{x}}_t \in [0,1]^{d}$. From this scaling, we use the sigmoid function in the last layer of the generator to make a generated sample $\widehat{\bm{x}}^*_t \in (0,1)^d$. 
As discussed in section \ref{sec:GAN}, it is important to keep $Supp(p_g) \simeq Supp(p_d)$ in the training of GAN. Using the sigmoid function and $\nu = 0$ implicitly regularizes the generator to guarantee $\bm{x}^*_t \in Supp(p_d)$. When the underlying distribution has an extremely sharp peak near $\max(x_i)$ or $\min(x_i)$, \emph{e.g.}, noiseless pendulum data, it is helpful to have $0 < \nu \ll | \max(x_i) - \min(x_i) |$ so that $Supp(p_g)$ includes $\max(x_i)$ and $\min(x_i)$. In this study, $\nu$ is set to zero. The detailed model structures are shown in Appendix \ref{sec:arch}. We find that a pre-training of $\Psi_G$ and $\Psi_D$ helps CR-GAN converge faster. The pre-training can be done by adding a linear layer to $\Psi_G$ (and $\Psi_D$) and train as a simple regression model with the standard mean-square loss function. After the pre-training, the linear layers are discarded. Once the whole training is completed, the data-driven forecast is performed by a Monte Carlo simulation \cite{Yeo21,Yeo19}. The Monte Carlos method is outlined in Appendix \ref{sec:MC}.

\subsection{Bi-Gaussian Auto-Regressive Process} \label{subsec:AR}

\begin{figure}
\centering
\includegraphics[width=0.95\textwidth]{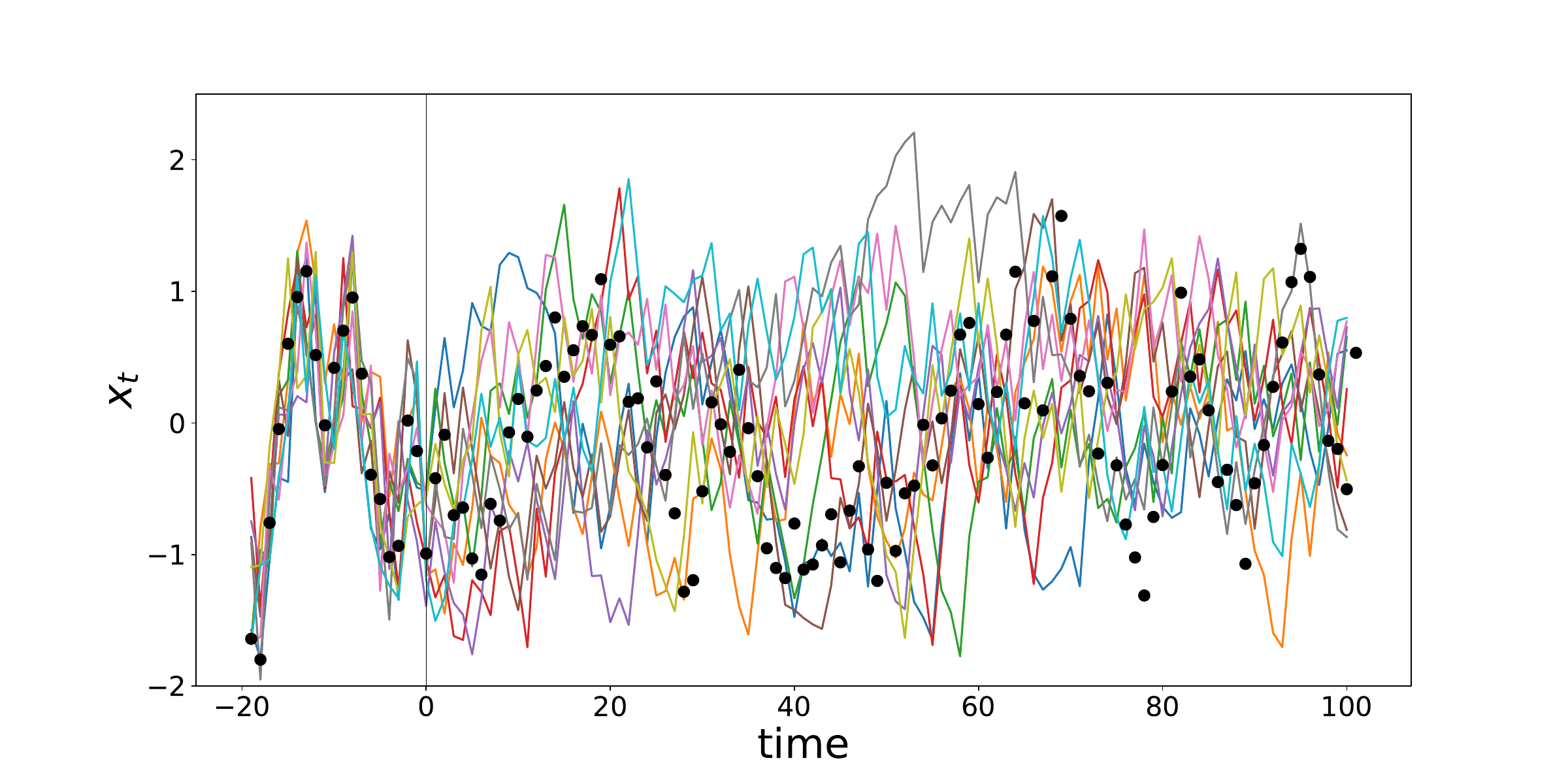}
\caption{Trajectories generated by CR-GAN (solid lines). The forecast starts at $t=0$. The solid circles denote the ground truth data.} \label{fig:binorm_traj}
\end{figure}

For the first numerical experiment, we consider an autoregressive process of order 1;
\begin{equation}
x_{t+1} = 0.8 x_t + \epsilon_t,
\end{equation}
where the noise process is
\begin{equation}
\epsilon_t \sim 0.5 \mathcal{N}(-2 \sigma,\sigma^2) + 0.5 \mathcal{N}(2\sigma,\sigma^2).
\end{equation}
The noise parameter $\sigma = 0.2$ is used. The length of the time series data used for the training is $4\times 10^4$. Because of the simplicity of the problem, we only consider CR-GAN to explore the effects of the first consistency condition (\ref{eqn:cond_marginal}). Fig. \ref{fig:binorm_traj} shows the trajectories generated by CR-GAN together with the ground truth data. The time series data is provided to CR-GAN for $t \in [-50,0]$ and CR-GAN makes predictions for $t \in [1,100]$, \emph{i.e.}, $\bm{x}^*_{1:100}$. 

\begin{figure}
\centering
\includegraphics[width=0.95\textwidth]{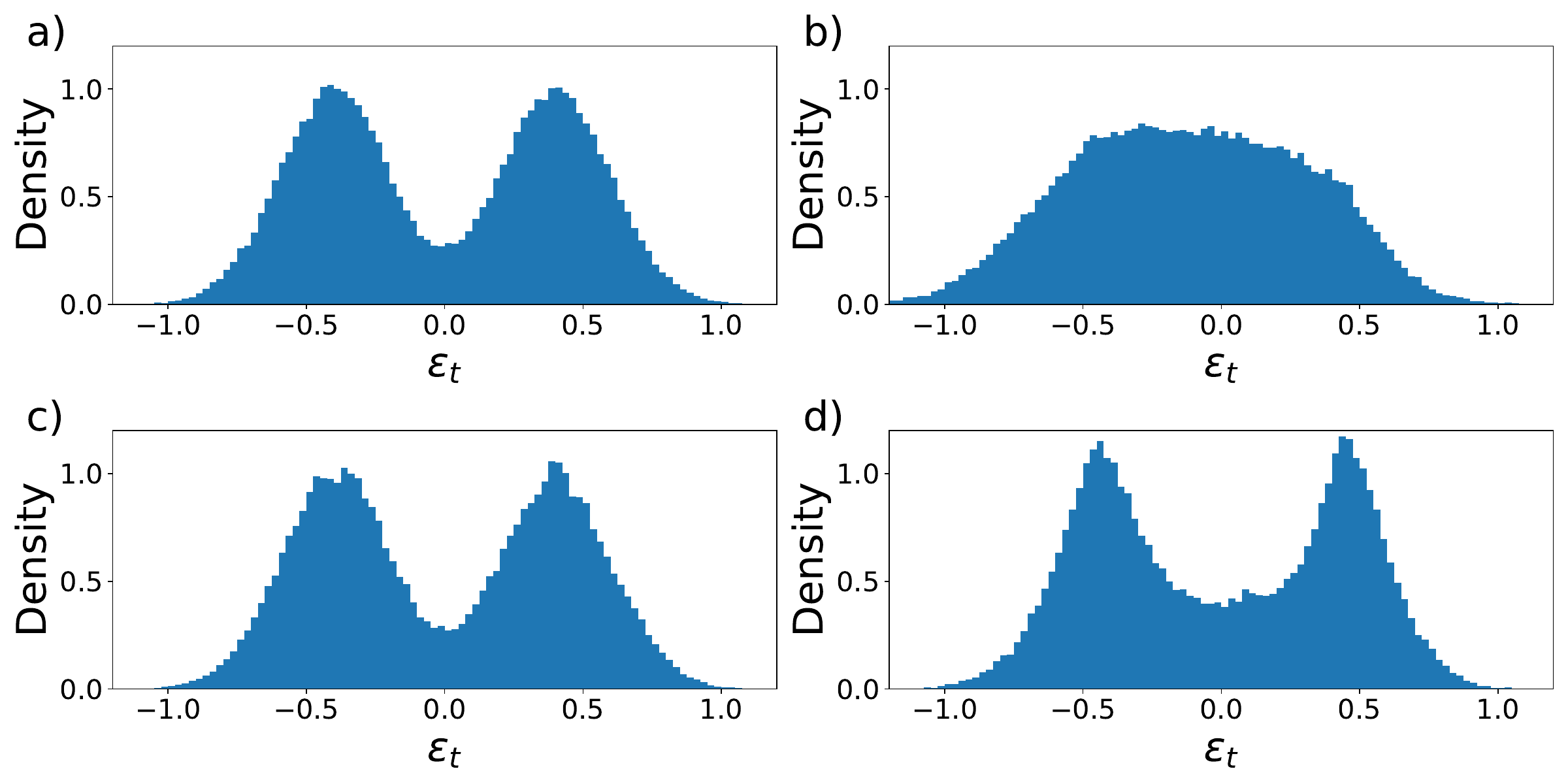}
\caption{Comparison of the estimated probability distribution of $\epsilon_t$. The histograms are computed from (a) data, (b) CR-GAN without any regularization, (c) CR-GAN with $\gamma = 0.2$, and (d) CR-GAN with $\gamma = 1.6$. The regularization parameter is fixed, $\lambda = 100$.} \label{fig:binorm_comp_dist}
\end{figure}

Fig. \ref{fig:binorm_comp_dist} shows the estimated noise from the generated samples, $\bm{x}^*_{1:100}$. A Monte Carlo simulation with 1,000 samples is performed to compute $x^*_{1:100}$. The histograms in Fig. \ref{fig:binorm_comp_dist} are evaluated by estimating the noise as
\begin{equation}
\epsilon^*_t = x^*_{t+1} - 0.8 x^*_t~\text{for}~t=1,\cdots,99.
\end{equation}
In Fig. \ref{fig:binorm_comp_dist} (c), It is shown that CR-GAN makes a very good approximation of the ground truth distribution (Fig. \ref{fig:binorm_comp_dist} a), compared to the GAN model without the consistency regularization (Fig. \ref{fig:binorm_comp_dist} b). As shown in (\ref{eqn:mmd_kernel}), the MMD depends on a scale parameter of the Gaussian kernel, $\gamma$. Fig. \ref{fig:binorm_comp_dist} (c) and (d) show the effects of $\gamma$ on the estimation of the probability distribution. At $\gamma = 1.6$ (Fig. \ref{fig:binorm_comp_dist} d), although the estimated probability distribution has a bi-modal structure, the overall shape becomes noticeably different from the ground truth. 

\begin{table}
\center{
\caption{Kullback-Leibler divergence for CR-GAN trained with the regularization coefficient, $\lambda = 100$. Here, $\gamma = \infty$ indicates CR-GAN without the MMD regularization.} \label{tbl:binormal_kl}
\begin{tabular}{c|ccc}
\hline \hline
$\gamma$ & 0.2 & 1.6 & $\infty$ \\
\hline
$D_{KL}$ & 0.0011 & 0.0178 & 0.1425\\
\hline \hline
\end{tabular}
}
\end{table}

To make a quantitative comparison, the Kullback-Leibler (KL) divergence is computed for discrete probability distributions as
\begin{equation}
D_{KL}(\bm{Q}\|\bm{P}) = \sum_{i=1}^K Q_i \log \frac{Q_i}{P_i},
\end{equation}
where $\bm{P}$ and $\bm{Q}$ denote the empirical probability distributions for the data and the generated samples, respectively. The KL divergence is computed in the interval, $(-1.3,1.3]$ with the bin size of $\delta = 0.05$. It is shown that, when $\gamma = 0.2$, the KL divergence is less than 1\% of $D_{KL}$ of the GAN without the MMD regularization. The KL divergence becomes larger for $\gamma = 1.6$, but it is still about 15\% of $D_{KL}$ without the MMD regularization.

\begin{figure}
\includegraphics[width=0.99\textwidth]{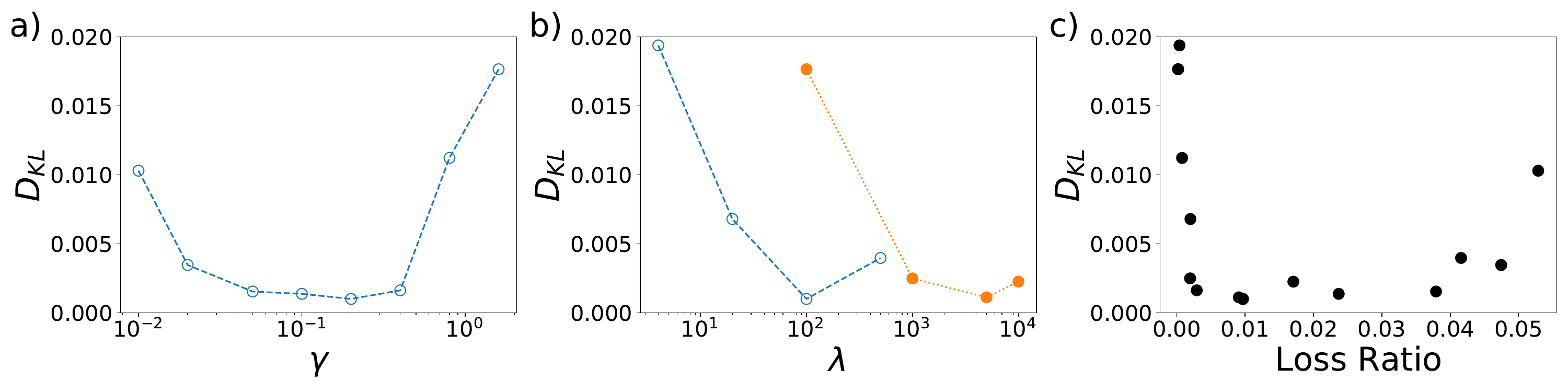}
\caption{Effects of the MMD parameters. (a) Effects of the scale parameter, $\gamma$, for a fixed regularization coefficient $\lambda = 100$. (b) Effects of $\lambda$ for $\gamma = 0.2$ ({\color{blue}$\circ$}) and $1.6$ ({\color{orange}$\bullet$}). (c) $D_{KL}$ as a function of the loss ratio.  } \label{fig:binormal_kl}
\end{figure}

There are two parameters in the MMD regularization for the first consistency condition; the regularization coefficient, $\lambda$, and the scale parameter of the MMD kernel, $\gamma$. To investigate the effects of $\gamma$, we fix $\lambda = 100$ and calculated $D_{KL}$ for a range of $\gamma$ (Fig. \ref{fig:binormal_kl} a). It is shown that $D_{KL}$ remains almost constant for $0.05 \le \gamma \le 0.4$ and then starts to increase. In Fig. \ref{fig:binormal_kl} (b), now the effects of $\lambda$ are shown for $\gamma = 0.2$ and 1.6.  For a fixed $\gamma$, $D_{KL}$ first decreases and then starts to increase, as $\lambda$ increases. To achieve the similar $D_{KL}$, $\lambda$ for $\gamma = 1.6$ needs to be about two orders of magnitude larger than $\lambda$ for $\gamma = 0.2$. The huge difference in the optimal $\lambda$ will make it difficult to tune the regularization parameter in practice. 

In Fig. \ref{fig:binormal_kl} (c), it is shown that the results in Fig. \ref{fig:binormal_kl} (a -- b) are collapsed into one curve, when scaled by the ratio between the loss function and the MMD regularization. Define ``Loss ratio'' as
\begin{equation}
\text{Loss ratio} = \frac{\mathcal{R}_M}{\frac{1}{T}\sum_{i=1}^T \mathcal{L}_t},
\end{equation}
where $\mathcal{R}_M$ is the MMD regularization defined in (\ref{eqn:r_m}). The Loss ratio is computed by the average over the last 5000 iterations of the SGD solver. It is shown that $D_{KL}$ stays almost flat when the Loss ratio is in the range of $(0.002,0.036)$. When the Loss ratio is too small ($<0.002$), the MMD regularization does not play a significant role in the optimization problem, which makes $D_{KL}$ larger. On the other hand, when the Loss ratio becomes too large ($>0.04$), the generator tries to satisfy the marginal distribution, instead of learning the conditional distribution, which again makes $D_{KL}$ grow. Note that Fig. \ref{fig:binormal_kl} (c) encompasses $\gamma \in [0.01,1.6]$ and $\lambda \in [4,10^{4}]$. The results suggest that, instead of trying to find the optimal $\gamma$ and $\lambda$, it is more important to tune the MMD regularization parameters such that the relative magnitude between the MMD regularization and the generator loss function is within the range shown in Fig. \ref{fig:binormal_kl} (c). 

\subsection{Mackey-Glass Time Series} \label{subsec:MG}

\begin{figure}
\centering
\includegraphics[width=0.99\textwidth]{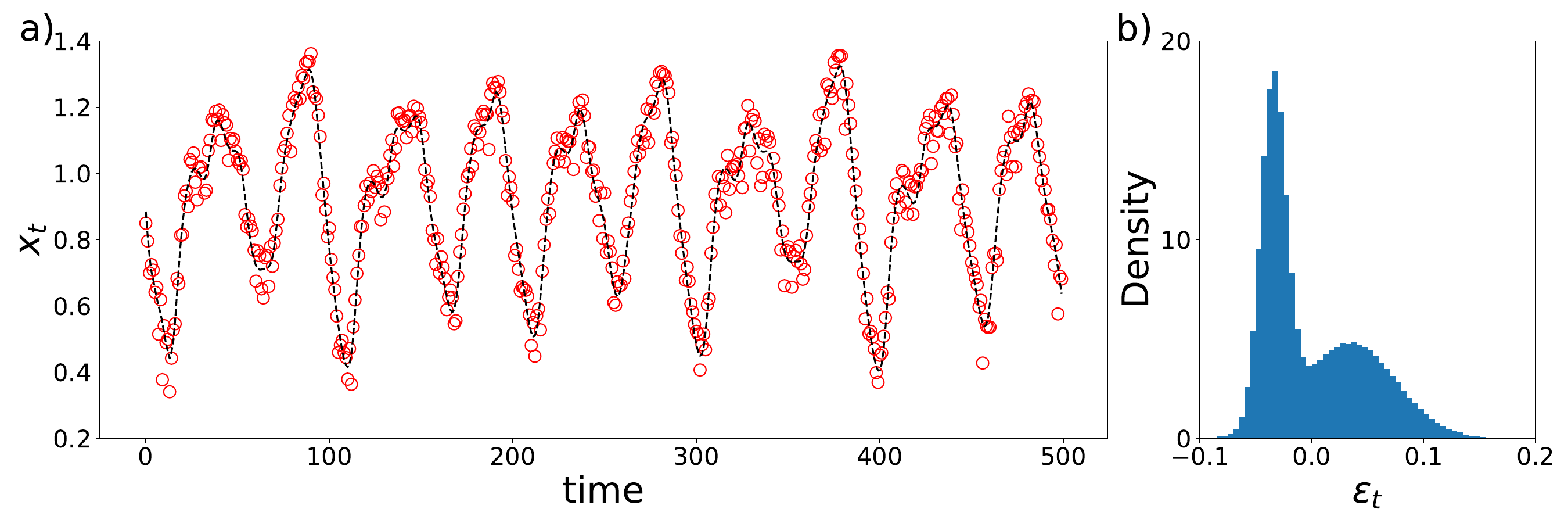}
\caption{Noisy observation of the Mackey-Glass time series. In (a), the solid line is the noiseless Mackey-Glass time series and the hollow circles denote the noisy observations, and (b) shows the probability distribution of the observation noise.} \label{fig:MG_traj}
\end{figure}

The Mackey-Glass dynamical system is a nonlinear dynamical system with time-delay dynamics \cite{Mackey77}. The equation for the Mackey-Glass dynamical system is
\begin{equation} \label{eqn:MG}
\frac{d \phi(t)}{dt} = \frac{a \phi(t-\tau)}{1+\phi^b(t-\tau)} - c \phi(t).
\end{equation}
The data is generated for the the parameters $(a,b,c,\tau) = (0.2,10,0.1,17)$, where the Mackey-Glass dynamical system becomes chaotic \cite{Farmer82,Gers01}. The time interval between consecutive data points is set to $\delta t = 1$. The time delay parameter ($\tau=17$) indicates that the time evolution at time step $t$ depends on the information at $t-\tau$, which is 17 time steps ago. After the noiseless data is generated by solving (\ref{eqn:MG}), noisy observations are made by adding white noise,
\begin{equation}
x_t = \phi(t_0 + t \delta t)+\epsilon_t,
\end{equation}
in which
\begin{equation}
\epsilon_t \sim 0.5 \mathcal{N}(3 \sigma,\sigma^2) + 0.5 \mathcal{N}(-3\sigma,13\sigma^2).
\end{equation}
Here, we use $\sigma = sd(\phi)\times 0.05$, in which $sd(\phi)$ denotes the standard deviation of the noiseless data, $\phi(t)$. The standard deviation of $\epsilon_t$ is 20\% of $sd(\phi)$. The time series data and the probability distribution of $\epsilon_t$ are shown in Fig. \ref{fig:MG_traj}. The length of the time series data to train the model is $4\times 10^5$. Five different models are trained; CR-GAN, CR-GAN-M$_1$, CR-GAN-M$_5$, CR-GAN-D$_1$, and CR-GAN-D$_5$. The MMD regularization coefficient of the first consistency condition, $p_g(\bm{x})$, is set to $\lambda_1 = 500$. The second MMD regularization coefficients for CR-GAN-M$_1$ and CR-GAN-M$_5$ are set to $\lambda_2=500$ and 200, respectively. Note that, contrary to CR-GAN-M$_n$, CR-GAN-D$_n$ does not have the second regularization coefficient, $\lambda_2$. For all the cases, the MMD kernel scale parameter is fixed, $\gamma = 0.2$.

\begin{figure}
\centering
\includegraphics[width=0.99\textwidth]{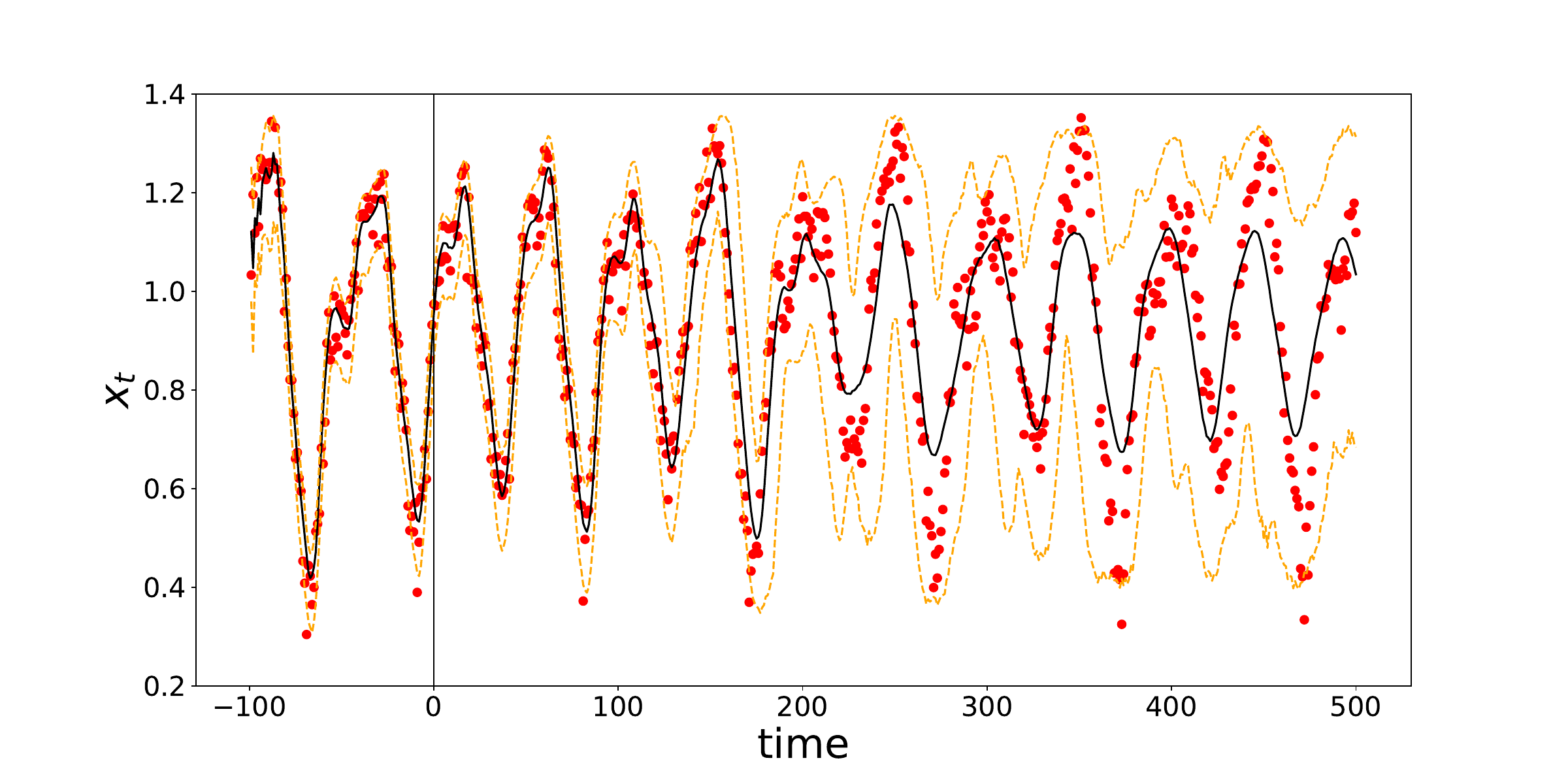}
\caption{500-step prediction of the noisy Mackey-Glass time series. The solid line and two dashed lines indicate the mean and 95\% prediction intervals, respectively. The time series data are shown as the solid circles. The prediction starts at $t=0$.} \label{fig:MG_pred}
\end{figure}

Fig. \ref{fig:MG_pred} shows a 500-step prediction of the noisy Mackey-Glass time series data by CR-GAN-D$_5$. The time series data is provided to CR-GAN-D$_5$ for $t = -100 \sim 0$, and a Monte Carlos simulation with 1,000 samples is performed to compute the mean and 95\% prediction intervals for $t = 1 \sim 500$.
The prediction interval $\text{PI}_t(p)$ for the coverage probability $p$ at time $t$ is defined as $\text{PI}_t(p) = [L_t(p),U_t(p)]$, where
\[
L_t(p) = \Phi_t^{-1}\left(\frac{1}{2}(1-p)\right), ~\text{and}~U_t(p) = \Phi_t^{-1}\left(\frac{1}{2}(1+p)\right),
\]
and $\Phi_t$ denote the empirical cumulative distribution at time $t$. Since the Mackey-Glass time series is chaotic, it is shown that, in general, the prediction interval widens as the forecast horizon increases. The mean of the prediction stays very close to the ground truth data up to $t=200$, then starts to deviate. As observed in the previous studies using RNN, the width of the prediction interval is not a monotonically increasing function of time \cite{Yeo19}. The width of the prediction interval may increase or decrease following the oscillatory pattern of the time series data.

For a quantitative study, we estimate the empirical coverage probabilities (ECP) for five prediction intervals, $\text{PI}(p)$ for $p \in \{0.6,0.7,0.8,0.9,0.95\}$. The empirical coverage probability is computed as
\begin{equation}
\text{ECP}(p) = \frac{1}{N H}\sum_{i=1}^N \sum_{t=1}^H \chi_{\text{PI}^i_t(p)}(x^{* (i)}_t).
\end{equation}
Here, $\text{PI}^i_t(p)$ is the prediction interval with the coverage probability $p$ at time step $t$ for the $i$-th trajectory, $\chi_A(x)$ is an indicator function, which is one if $x \in A$ and zero otherwise. In other words, $\text{ECP}(p)$ is estimated by computing the fraction of the ground truth data covered by $\text{PI}(p)$. The prediction interval, $\text{PI}^i_t(p)$, is evaluated by a Monte Carlo simulation with $1,000$ samples. The total number of trajectories to compute ECP is $N=500$ and the prediction horizon is set to $H = 1000$. The difference, $\text{ECP}(p) - p$, provides a quantitative measure for an error in the predictive probability distribution.

\begin{table}
\center{
\caption{Empirical Coverage Probability (ECP) for $1000$-step prediction of the noisy observation of the Mackey-Glass time series. SAD indicates the Sum of Absolute Difference between ECP and $p$.} \label{tbl:MG_Emp_Cov}
\begin{tabular}{c|cccccc}
\hline \hline
$p$ & RNN & CR-GAN & $\text{CR-GAN-M}_1$ &  $\text{CR-GAN-G}_1$ & $\text{CR-GAN-M}_5$ &$\text{CR-GAN-G}_5$ \\
\hline
0.60 & 0.659 & 0.634 & 0.566 & 0.619  & 0.608 & 0.599\\
0.70 & 0.760 & 0.731 & 0.662 & 0.712 & 0.702 & 0.693\\
0.80 & 0.854 & 0.825 & 0.763 & 0.802 & 0.797 & 0.791\\
0.90 & 0.936 & 0.915 & 0.870 & 0.894 & 0.892 & 0.890\\
0.95 & 0.973 & 0.958 & 0.928 & 0.942 & 0.943 & 0.942\\
\hline
SAD & 0.232 & 0.113 & 0.161 & 0.047 & 0.028 & 0.035\\
\hline \hline
\end{tabular}
}
\end{table}

The empirical coverage probability is shown in Table \ref{tbl:MG_Emp_Cov}. Here, RNN refers to a standard RNN model with the Gaussian distribution assumption. A two-layer Gated Recurrent Unit with 128 hidden states is used for the RNN, which is the same RNN structure used in CR-GAN models. It is shown that ECPs of RNN are always larger than $p$. At $p=0.6$ and 0.7, there are about 6\% difference between PI and ECP. CR-GAN shows a better performance compared to RNN. The sum of absolute difference,
\[
\text{SAD} = \sum_p |\text{ECP}(p) - p|,~\text{for}~p \in \{0.6,0.7,0.8,0.9,0.95\},
\]
of CR-GAN is less than half of SAD of RNN. Still, CR-GAN shows about 3\% difference at $p = 0.6$ and 0.7. When the second consistency condition is imposed by the first-order temporal difference, $p(\Delta_1 x_t)$, CR-GAN-M$_1$ does not show noticeable improvement over CR-GAN in terms of SAD, while SAD of CR-GAN-G$_1$ is less than half of CR-GAN.
When the fifth-order temporal difference, $p(\Delta_5 x_t),$ is used for the regularization, CR-GAN-M$_5$ and CR-GAN-D$_5$, SAD is reduced further. While CR-GAN-M$_5$ shows the best performance in terms of SAD, the difference between CR-GAN-M$_5$ and CR-GAN-D$_5$ is only marginal. It should be noted that, while CR-GAN-M$_5$ requires a careful tuning of both $\lambda_1$ and $\lambda_2$, CR-GAN-D$_5$ has one less tuning parameter, only $\lambda_1$, which makes it easier to tune. It is also found that using a higher order temporal difference, \emph{e.g.}, $p(\Delta_{10} x_t)$, does not improve the prediction accuracy. SADs of CR-GAN-M$_{10}$ and CR-GAN-D$_{10}$ are 0.041 and 0.072, respectively.

\subsection{Lorenz System} \label{subsec:Lorenz}

We now investigate the behaviors of CR-GAN by using the random Lorenz system \cite{Gradisek00};
\begin{equation}
\frac{d}{dt}
\begin{bmatrix}
x \\ y \\ z
\end{bmatrix}
=
\begin{bmatrix}
10(y - x) \\ x(28-z)-y \\ xy - \frac{8}{3}z
\end{bmatrix}
+
\begin{bmatrix}
4 & 5 & 3 \\
5 & 5 & 6 \\
3 & 6 & 10
\end{bmatrix}
\begin{bmatrix}
\eta_x(t) \\ \eta_y(t) \\ \eta_z(t)
\end{bmatrix},
\end{equation}
in which $\eta_i$ denotes the Gaussian white noise. The sampling interval of the time series data is $\delta t = 0.01$. 
The length of the time series data to train the model is $4\times 10^5$. For the random Lorenz system, we mainly consider CR-GAN, CR-GAN-D$_1$ and CR-GAN-D$_5$, because, while CR-GAN-M$_n$ shows a similar performance with CR-GAN-D$_n$, it is difficult to tune the two free parameters, $\lambda_1$ and $\lambda_2$. In the training, $\lambda_1 = 500$ and $\gamma = 0.2$ are used for all the models.

\begin{figure}
\centering
\includegraphics[width=0.99\textwidth]{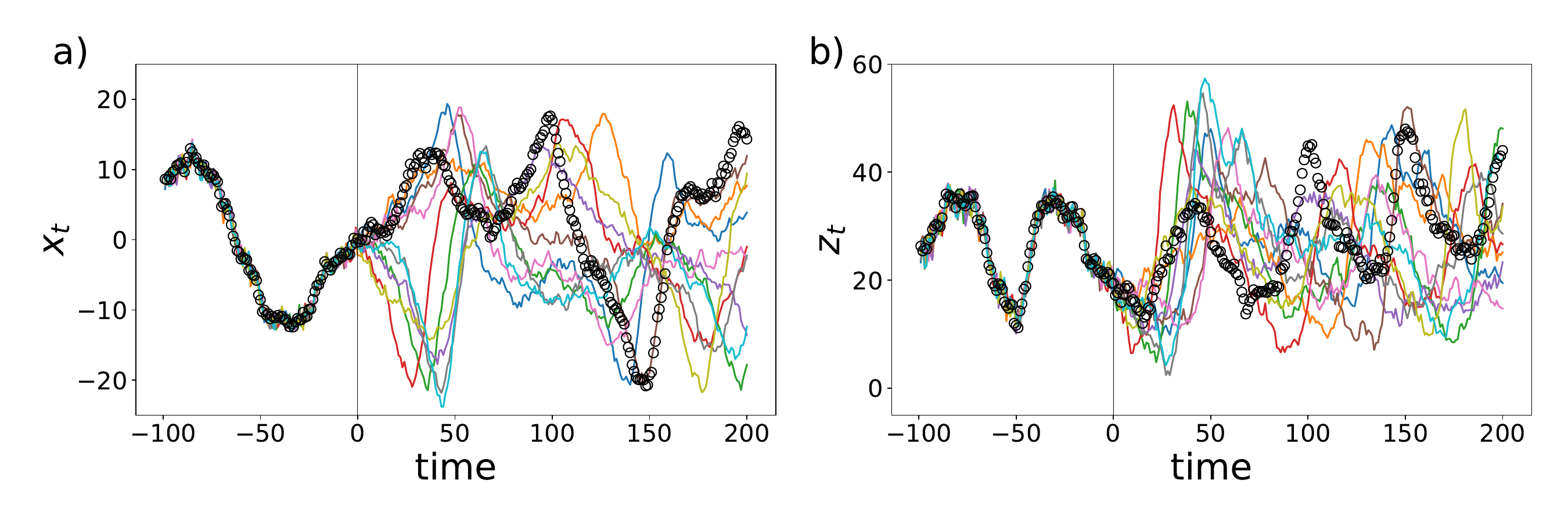}
\caption{Monte Carlo samples for the prediction of the random Lorenz system. The hollow circles denote the ground truth data, and the lines are sample trajectories from CR-GAN-D$_5$. The horizontal axes denote the time steps.}\label{fig:Lorenz_sample}
\end{figure}

Fig. \ref{fig:Lorenz_sample} shows ten trajectories from the Monte Carlo simulation of CR-GAN-D$_5$. The time series data for $t\in[-100,0]$ are used to make a prediction of the next 200 time steps. In Fig. \ref{fig:Lorenz_sample} (a), it is shown that there is a bifurcation of the trajectories of $x_t$ right after the prediction begins. On the other hand, the trajectories of $z_t$ show oscillatory patterns similar to the ground truth data.

\begin{figure}
\centering
\includegraphics[width=0.7\textwidth]{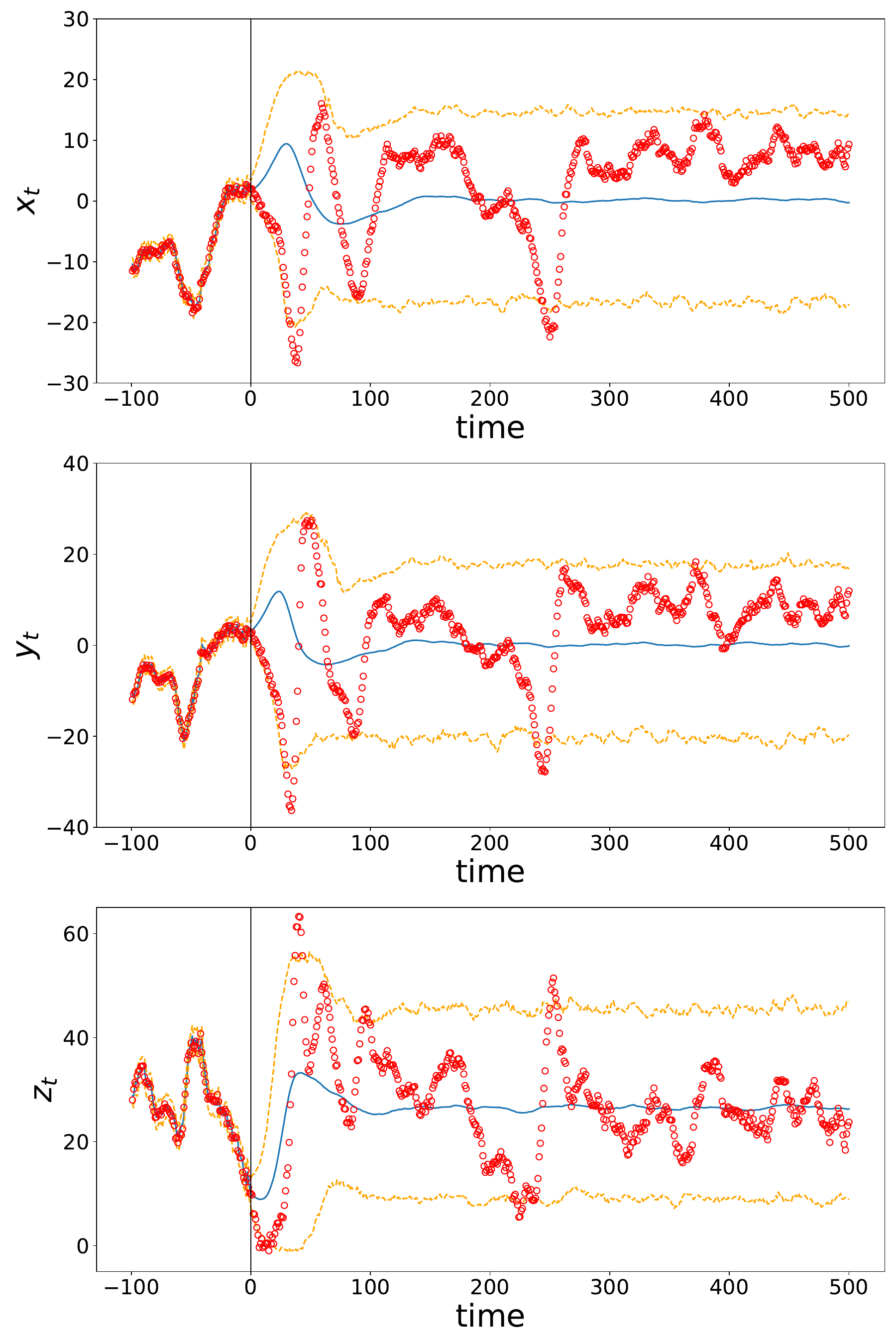}
\caption{500-step ahead prediction of the random Lorenz system by using CR-GAN-D$_5$. The solid line and two dashed lines indicate the mean and 95\% prediction intervals, respectively. The time series data are shown as the solid circles. The prediction starts at $t=0$. The horizontal axes denote the time steps.} \label{fig:Lorenz_pred}
\end{figure}

Fig. \ref{fig:Lorenz_pred} shows a 500-step prediction of the random Lorenz system by using CR-GAN-D$_5$. The prediction intervals of both $x_t$ and $y_t$ rapidly increase from the beginning of the prediction, which is consistent with the bifurcation observed in Fig. \ref{fig:Lorenz_sample} (a). For all the components of the Lorenz system ($x_t,y_t,z_t$), the 95\% prediction intervals first increase and then decrease before reaching a stationary state for $t > 150$. 

The average empirical coverage probability (AECP) is shown in Table \ref{tbl:Lorenz_Emp_Cov}. Here, AECP is computed by an average of ECPs for $x_t$, $y_t$, and $z_t$. Again, it is shown that CR-GAN models have much smaller AECP compared to RNN. Although AECP of CR-GAN-G$_5$ is smaller than that of CR-GAN-G$_1$, the difference is only marginal. CR-GAN-M$_5$ is trained with $\lambda_2 = 50$ and $\gamma = 0.2$. It is shown that AECP of CR-GAN-M$_5$ is much larger than that of CR-GAN-D$_1$, CR-GAN-D$_5$, or even CR-GAN, which may be due to a suboptimal choice of the regularization coefficients, $\lambda_1$ and $\lambda_2$. Because of the interaction of the two MMD regularizations in CR-GAN-M$_n$, it is difficult to find an optimal $\lambda_1$ and $\lambda_2$. On the other hand, CR-GAN-D$_n$ has only one regularization coefficient, $\lambda_1$, which can be chosen, as suggested in Fig. \ref{fig:binormal_kl} (c). 

\begin{table}
\center{
\caption{Average Empirical Coverage Probability. SAD indicates the Sum of Absolute Difference.} \label{tbl:Lorenz_Emp_Cov}
\begin{tabular}{c|ccccc}
\hline \hline
$p$ & RNN & CR-GAN & $\text{CR-GAN-G}_1$ & $\text{CR-GAN-G}_5$ & $\text{CR-GAN-M}_5$\\
\hline
0.60 & 0.568 & 0.602 & 0.599 & 0.603 & 0.619 \\
0.70 & 0.671 & 0.707 & 0.704 & 0.701 & 0.721\\
0.80 & 0.770 & 0.812 & 0.805 & 0.800 & 0.824\\
0.90 & 0.867 & 0.908 & 0.902 & 0.898 & 0.918\\
0.95 & 0.919 & 0.954 & 0.949 & 0.946 & 0.955\\
\hline
SAD & 0.155 & 0.033 & 0.013 & 0.010 & 0.087\\
\hline \hline
\end{tabular}
}
\end{table}

\begin{figure}
\includegraphics[width=0.99\textwidth]{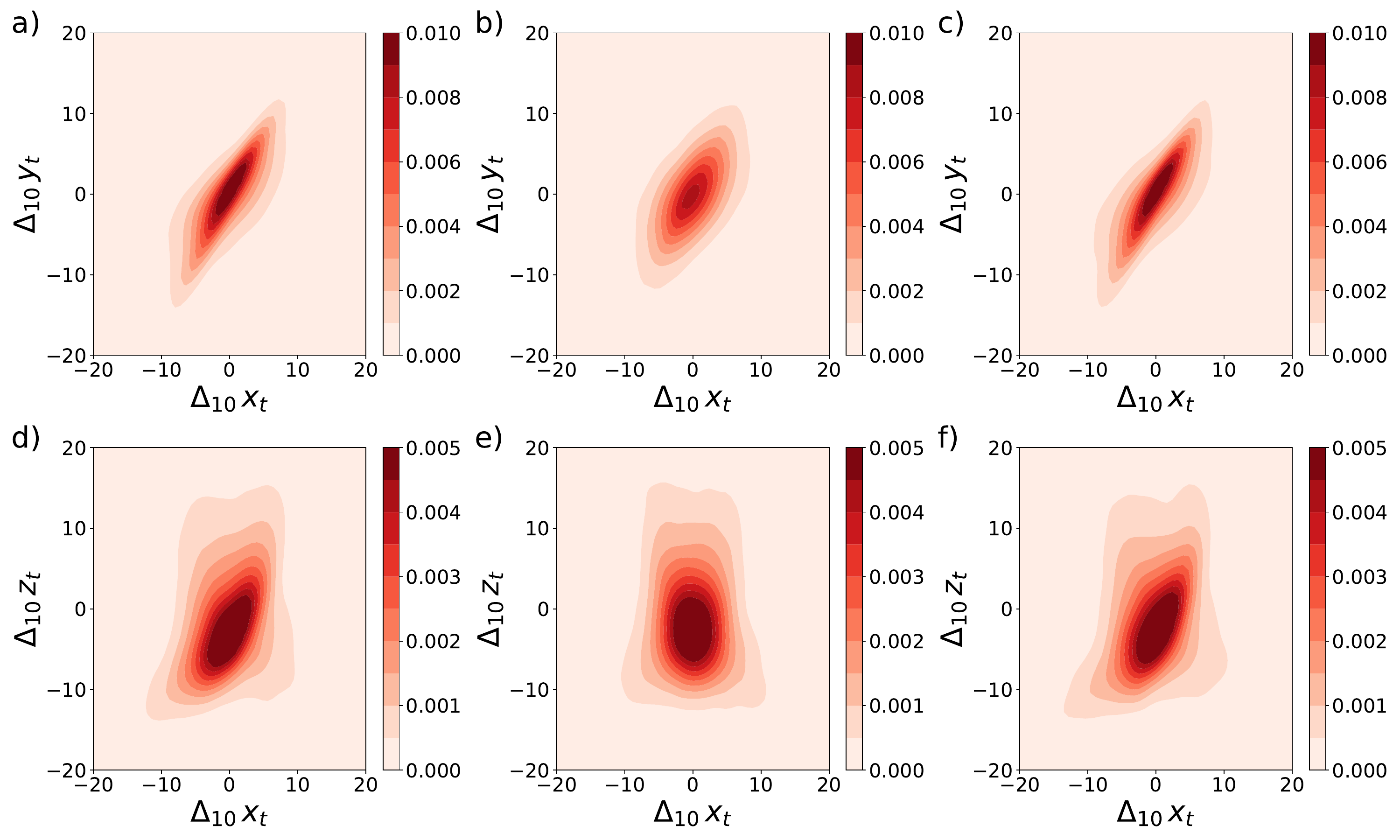}
\caption{ Two-dimensional joint probability density functions of the 10-th order temporal difference, $p(\Delta_{10} \bm{x}_t)$, from the ground truth data (a,d), RNN (b,e), and CR-GAN-G$_5$ (c,f). The color contours denote the probability density estimated by a kernel density estimation.}\label{fig:Lorenz_density}
\end{figure}

The empirical coverage probability is computed based on the marginal distribution of each component, \emph{e.g.}, $p(x_{t+n}|x_{0:t},y_{0:t},z_{0:t})$. To demonstrate the capability of CR-GAN in the estimation of the joint probability distribution, joint probability distributions between two components are shown in Fig. \ref{fig:Lorenz_density}. The joint probability distributions are computed for the 10-th order temporal difference, \emph{e.g.}, $p(\Delta_{10} x_t, \Delta_{10} y_t)$, of the generated trajectories. Total 1,000 trajectories are generated from different initial sequences. For each trajectory, the ground truth trajectory is used for the first 100 time steps, and CR-GAN-D$_5$ generates the next 500 time steps. A two-dimensional kernel density estimation with a Gaussian kernel is used to compute the joint probability distribution from randomly selected 50,000 samples. It is shown that the joint probability distributions from CR-GAN-D$_5$ are very close to the ground truth distribution, while the joint probability distributions from the standard RNN are noticeably different.
 
\begin{figure}
\centering
\includegraphics[width=0.8\textwidth]{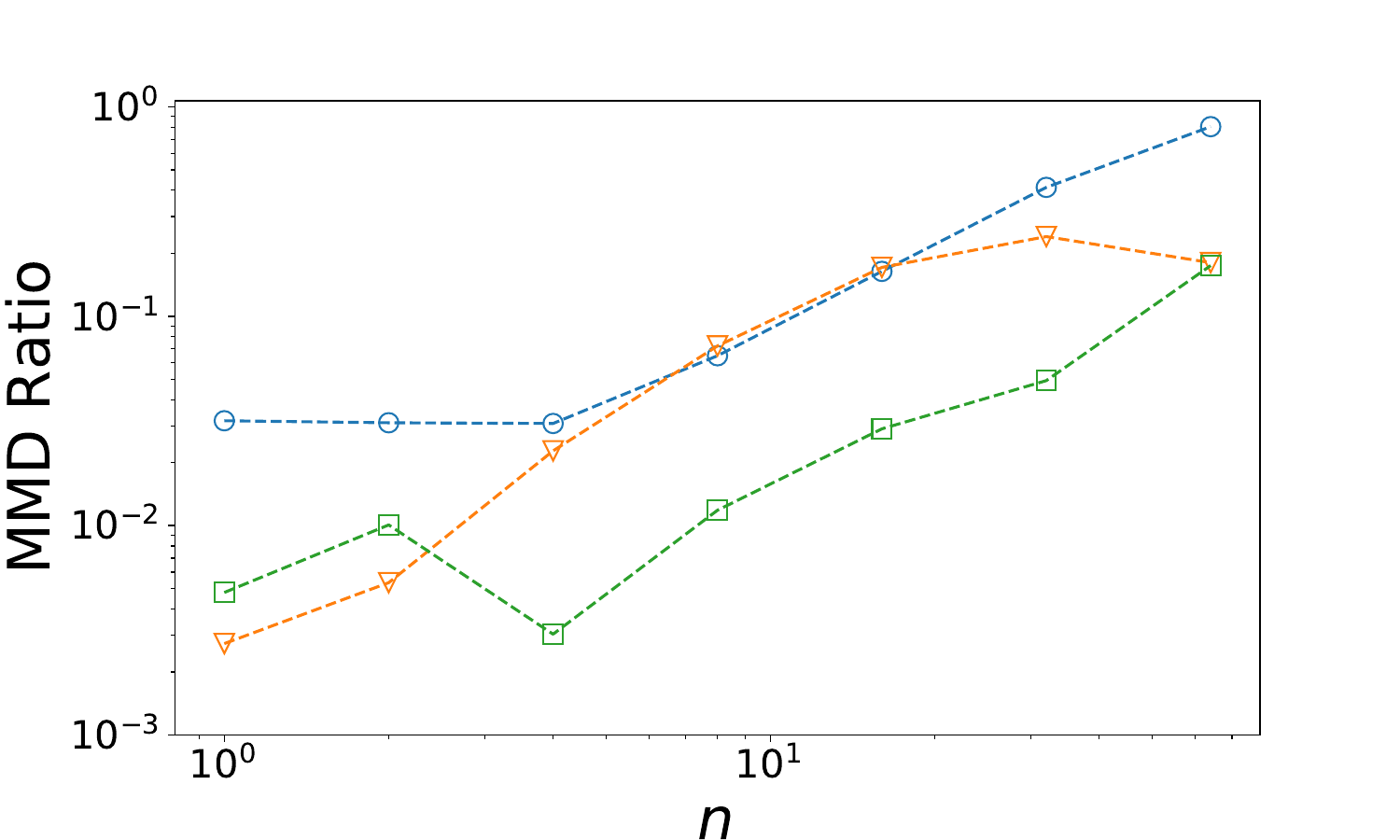}
\caption{MMD Ratios of CR-GAN (${\color{blue}\bigcirc}$), CR-GAN-D$_1$ (${\color{orange}\bigtriangledown}$), and CR-GAN-D$_5$ (${\color{green}\Box}$) as a function of the order ($n$) of the temporal difference, $\Delta_n \bm{x}_t$.} \label{fig:Lorenz_mmd}
\end{figure}

The maximum mean discrepancy is used to make a quantitative comparison about the joint probability distributions between different CR-GAN models. The Gaussian kernel is used for the computation of MMD,
\begin{equation}
k(\bm{u},\bm{u}^*) = \exp\left\{ \sum_{i=1}^d -\frac{(u_i-u^*_i)^2}{2\, \sigma_i^2} \right\},
\end{equation}
in which $\bm{u}$ and $\bm{u}^*$ denote $d$-dimensional random variables and $\sigma_i$ is the standard deviation of $u_i$. Here, the standard deviations for $(x_t,y_t,z_t)$ are computed from the data. Because the value of MMD is difficult to interpret, a ratio of MMD values between CR-GAN models and RNN is investigated,
\[
\Lambda = \frac{\text{MMD(CR-GAN)}}{\text{MMD(RNN)}}.
\]
Here, $\Lambda$ can be thought as a relative error. The same data to make Fig \ref{fig:Lorenz_density} are used to compute $\Lambda$. Fig. \ref{fig:Lorenz_mmd} shows $\Lambda$ with respect to the order of the temporal difference. The order of the temporal difference is varied from $n=1$ to $n=64$. It is shown that, when the second consistency condition is not imposed (CR-GAN), $\Lambda$ grows rapidly as $n$ increases. At $n=64$, $\Lambda$ of CR-GAN becomes larger than 0.8. On the other hand, CR-GAN-D$_5$ shows only a mild growth of $\Lambda$. Although the second consistency condition is imposed only for the 5-th order temporal difference, $\Lambda$ for $n=32$ still remains less than 0.05. CR-GAN-D$_1$ initially shows similar $\Lambda$ with CR-GAN-D$_5$, which increases rapidly for $4 \le n \le 16$. However, contrary to CR-GAN, $\Lambda$ of CR-GAN-D$_1$ does not increase for for $n \ge 32$. At $n=64$, $\Lambda$'s of CR-GAN-D$_1$ and CR-GAN-D$_5$ become roughly the same.

\section{Summary}  \label{sec:summary}
The consistency-regularized generative adversarial network (CR-GAN) is proposed for the data-driven simulation of random dynamical system. The CR-GAN consists of a RNN and a feedforward neural network. The RNN aims to learn the time marching operator and the feedforward neural network, which takes a white noise as one of the input variables, aims to learn the joint probability distribution of the stochastic process without a distributional assumption and draw a sample from the learned probability distribution.
Note that the structure of our GAN is similar to the conditional GAN proposed in \cite{Mirza14}, because one of the goals of CR-GAN is also to learn a conditional distribution. However, CR-GAN is combined with a RNN to learn the time marching structure and trained with a new objective function (\ref{eqn:GAN_loss_new}-\ref{eqn:GAN_loss_new_2}), which is designed to learn the predictive conditional distribution, $p(\bm{x}_{t+1}|\bm{x}_{0:t})$. Although GAN provides a flexible framework to learn a complex, high-dimensional joint probability distribution, training a GAN model is notoriously difficult. It is important to note that GAN is not free from the curse of dimensionality, which requires an exponential growth in the number of training samples as the dimension of the problem increases. While many of the previous studies focus on regularizing the behavior of the discriminator to facilitate the training, we propose a regularization strategy for the generator, based on the characteristics of the time series problem. 

The consistency-regularized generative adversarial network aims to learn the conditional distribution of a time series problem, $p(\bm{x}_{t+1}|\bm{x}_{0:t})$. The regularization for CR-GAN is based on integral constraints, which relates the conditional to the marginal distributions. The first consistency condition enforces a stationary state condition,
\[
\int p_g(\bm{x}_{t+1}|\bm{x}_{0:t})p_d(\bm{x}_{0:t}) d\bm{x}_{0:t} ~\simeq~ p_d(\bm{x}_{t+1}),
\]
by using a two sample test, called the Maximum Mean Discrepancy. The second consistency condition is from the stationarity of the predictive distribution,
\[
\int p_g(\Delta_n \bm{x}_t|\bm{x}_{0:t})p_g(\bm{x}_{t_0+1:t}|\bm{x}_{0:t_0}) p_d(\bm{x}_{0:t_0}) d\bm{x}_{0:t} ~\simeq~ p_d(\Delta_n \bm{x}_t),
\]
in which $\Delta_n \bm{x}_t = \bm{x}_{t+n} - \bm{x}_t$, and $1 < t_0 < t$ is the length of an initial spin-up time for CR-GAN. We propose to use either MMD (CR-GAN-M$_n$) or a discriminator (CR-GAN-D$_n$) to impose the second consistency condition.

The behaviors of CR-GAN are investigated by using three stochastic processes with varying complexities. First, CR-GAN is used to learn an auto-regressive process of order one, of which noise process is given by a bi-Gaussian distribution. It is clearly shown that the first consistency condition improves the accuracy of CR-GAN in learning the bi-Gaussian noise. In CR-GAN, there are two tuning parameters, one is the scale parameter of the MMD kernel, $\gamma$, and the other is the magnitude of the MMD regularization, $\lambda_1$. In the numerical experiments, it is shown that the ratio of the MMD regularization to the generator loss is more important parameter in the accuracy of CR-GAN, than the individual values of $\gamma$ and $\lambda_1$. It is shown that the Kullback-Leibler divergence between the generator distribution of CR-GAN and the data distribution stays almost constant when the ratio between the MMD regularization and the generator loss is between 0.002 and 0.4, which increases for a smaller ($< 0.002$) or larger ($>0.4$) values of the Loss ratio. This result provides a guideline to tune the hyperparameters. In the second numerical experiments on the noisy observation of the chaotic Mackey-Glass time series, it is shown that the second consistency condition does improve the accuracy of the multiple-step prediction. 
While there is virtually no difference in the accuracy between CR-GAN-M$_n$ and CR-GAN-D$_n$, CR-GAN-D$_n$ is easier to train because it has one less free parameter than CR-GAN-M$_n$. It is worthwhile to note that CR-GAN-D$_n$ is more computationally intensive during the training due to the evaluation of the additional discriminators. However, once trained, the computational time for the simulation is identical. Finally, CR-GAN is tested against a random Lorenz system. It is shown that CR-GAN-D$_n$ makes a reliable estimation of the joint probability distribution of $\Delta_n \bm{x}_t$.
Even though CR-GAN-D$_5$ is trained with only the fifth order temporal difference, the error in the joint probability distribution does not increase noticeably up to $\Delta_{32}\bm{x}_t$, while CR-GAN, which trained with only the first consistency condition, shows a rapid increase in the error as the order of the temporal difference increases.

In the present study, we mainly focus on learning from a multivariate time series data. Although the problem is formulated and tested for dynamical systems with one or zero time-delay parameter, the method can be directly applied to a broader class of problems, e.g., multiple time delays or integral operators. Since GAN aims to directly learn the data generating distribution without a distributional assumption, it requires a large number of training data. In all of our numerical experiments, the size of the data set is fixed, $T=4\times 10^5$. The consistency regularizations of CR-GAN, CR-GAN-M$_n$, and CR-GAN-D$_n$ facilitate the training by imposing constraints on marginal distributions, which have much lower dimensions than the predictive conditional distributions. The effects of the data size on the accuracy of GAN and the effects of the regularizations to mitigate the dependency on the data size are subjects of further investigation. For a more complex spatio-temporal process, such as geophysical flows, training a GAN will suffer from the same difficulties with the multivariate time series, largely due to the finite size of the training examples. We expect that our regularization strategy can also be applied to such kind of complex problems. Recently, a few promising generative models for time series problems have been proposed by using a normalizing flow \cite{Rasul21} or a Wasserstein auto-encoder \cite{Han21}. We believe that the proposed consistency regularization can be applied for those models to improve the stability in the training.

\section{Acknowledgement}
This work was supported by the Rensselaer-IBM AI Research Collaboration (http://airc.rpi.edu), part of the IBM AI Horizons Network (http://ibm.biz/AIHorizons).

\clearpage

\appendix

\section{Model Structure} \label{sec:arch}

\begin{figure}
	\centering
	\includegraphics[width=0.6\textwidth]{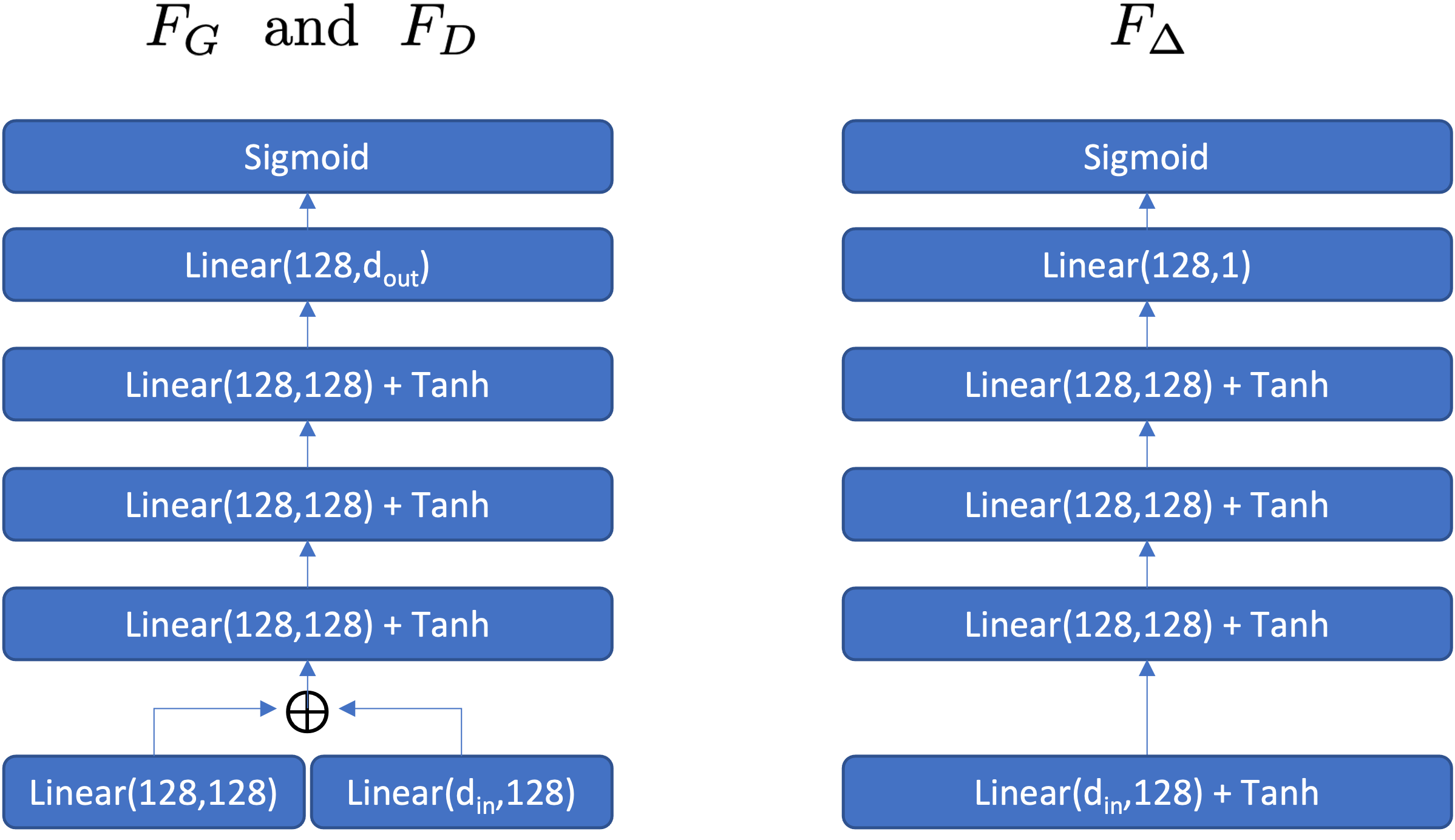}
	\caption{Sketches of the Generator($F_G$) and Discriminator ($F_D$) for $\bm{x}_t$ (left) and the Marginal Discriminator $F_\Delta$ for the temporal difference $\Delta_n \bm{x}_t$ (right). $d_{in}$ and $d_{out}$ denote the input and output dimensions. $d_{in}$ of $F_G$ is $\dim(\bm{z})$, while $F_D$ and $F_\Delta$ have the same $d_{in} = \dim(\bm{x}_t)$. The output dimension of $F_G$ is $d_{out} = \dim(\bm{x}_t)$ and that of $F_D$ is $d_{out} = 1$. In the sketch, Linear(a,b) denotes a linear layer, of which input and output dimensions are ``a'' and ``b'', respectively.  } \label{fig:sketch_arch}
\end{figure}

In this study, we use a two-level gated recurrent unit (GRU) \cite{Cho14} for the recurrent neural networks for the generator ($\Psi_G$) and the discriminator ($\Psi_D$). The structure of GRU is shown in Appendix \ref{sec:gru}.The same dimension is used for the hidden states of $\Psi_G$ and $\Psi_D$: $\dim(\bm{h}^G_t) = \dim(\bm{h}^D_t) = 128$. As shown in Fig. \ref{fig:sketch}, the hidden states are used as one of the inputs to multilayer feedforward neural networks, $F_G$ and $F_D$. The structures of $F_D$ and $F_G$ are shown in the left panel of Fig. \ref{fig:sketch_arch}. Here, Linear(a,b) indicates a linear transformation defined as
\begin{equation}
\text{Linear(a,b)}: \bm{o} = \bm{W}\bm{x}+\bm{c},
\end{equation}
in which $\bm{W} \in \mathbb{R}^{b\times a}$ and $\bm{c} \in \mathbb{R}^{b\times 1}$ are the weight and bias matrices and $\bm{x} \in \mathbb{R}^a$ is an input variable. The right panel of Fig. \ref{fig:sketch_arch} shows the structure of the discriminator ($F_\Delta$) for the marginal distribution of the temporal difference, $\Delta_n \bm{x}_t$.

\section{Gated Recurrent Unit} \label{sec:gru}
The gated recurrent unit (GRU) is proposed by \cite{Cho14} as a simplified version of the Long Short-Term Memory Network (LSTM) \cite{Hochreiter97}. Let $\bm{x}_t \in \mathbb{R}^{N_x}$ be an input vector to a GRU and $\bm{h}_t \in \mathbb{R}^{N_h}$ be the state vector of the GRU at time step $t$. In GRU, first auxiliary variables are computed as
\begin{align}
\bm{p}_t &= \sigma_g(\bm{W}_{px} \bm{x}_t + \bm{W}_{ph}\bm{h}_t+\bm{B}_p), \nonumber \\
\bm{q}_t &= \sigma_g(\bm{W}_{qx} \bm{x}_t + \bm{W}_{qh}\bm{h}_t+\bm{B}_q), \nonumber \\
\bm{r}_t &= \tanh(\bm{W}_{rx}\bm{x}_t + \bm{W}_{rh}(\bm{q}_t \odot \bm{h}_t) + \bm{B}_r). \nonumber \\ 
\end{align}
Here, $\sigma_g$ is the Sigmoid function, $\bm{W}_k$ for $k\in(px,ph,qx,qh,rx,rh)$ denotes the weight matrix and $\bm{B}_k$ for $k\in(p,q,r)$ is the bias vector, and $\odot$ is the element-wise multiplication operator, \emph{i.e.}, the Hadamard product. The dimensions of the auxiliary variables are the same with the state vector, \emph{i.e.}, $\dim(\bm{p}_t)=\dim(\bm{q}_t)=\dim(\bm{r}_t)=N_h$. Once the auxiliary variables are computed, the state vector is updated as,
\begin{equation}
\bm{h}_{t+1} = (1-\bm{p}_t) \odot \bm{h}_t + \bm{p}_t \odot \bm{r}_t.
\end{equation}
In the numerical experiments for the dynamical systems considered in this study, the difference between LSTM and GRU is not noticeable, while GRU has a smaller number of parameters.

\section{Monte Carlo simulation of CR-GAN} \label{sec:MC}

\begin{algorithm}[!tbh]
	\caption{Monte Carlo simulation of CR-GAN}
	\label{alg:MC}
	\textbf{Input}: Data $\bm{x}_{0:t}$, Monte Carlo sample size ($N_s$), forecast horizon ($H$)\\
	\textbf{Output}: $N_s$ trajectories sampled from $p(\bm{x}_{t+1:t+H}|\bm{x}_{0:t})$.
	
	\begin{algorithmic}
		\STATE 1. Sequential update of $\bm{h}^{G}$ for $\bm{x}_{0:t}$:.
		\STATE Set $\bm{h}^G_0 = \bm{0}$
		\FOR{$i=1,t+1$}
		\STATE $\bm{h}^G_i = \Psi_G(\bm{h}^G_{i-1},\bm{x}_{i-1})$
		\ENDFOR
		\STATE 2. Duplicate $\bm{h}_{t+1}$: 	
		$\displaystyle
		\bm{h}_{t+1}^{G, (1)} = \cdots = \bm{h}_{t+1}^{G, (N_s)}=\bm{h}^G_{t+1}
		$
		\STATE 3. Monte Carlo simulation:\\		
		\FOR{$i=1,H$}	
		\FOR{$j=1,N_s$}
		\STATE Draw a sample from $p_z(\bm{z})$: $\bm{z} \sim p_z(\bm{z})$
		\STATE Generate a sample: \hspace{3em} $\bm{x}^{* (j)}_{t+i} = F_G(\bm{z},\bm{h}^{G,(j)}_{t+i})$
		\STATE Update RNN: \hspace{6em}$\displaystyle
		\bm{h}^{G,(j)}_{t+i+1} = \Psi_G(\bm{h}^{G,(j)}_{t+i},\bm{x}^{* (j)}_{t+i}) 
		$
		\ENDFOR
		\ENDFOR
	\end{algorithmic}
\end{algorithm}

\bibliographystyle{siam}
\bibliography{ref_rnn}
	
\end{document}